\documentclass[letterpaper]{article} 
\usepackage{aaai25}  
\usepackage{times}  
\usepackage{helvet}  
\usepackage{courier}  
\usepackage[hyphens]{url}  
\usepackage{graphicx} 
\usepackage{natbib}  
\usepackage{caption} 
\usepackage{algorithm}
\usepackage{algorithmic}

\usepackage{newfloat}
\usepackage{listings}
\usepackage{xcolor}
\usepackage{bibentry}
\usepackage{tikz}

\usepackage[T1]{fontenc}
\usepackage[utf8]{inputenc}

\usepackage{amsmath}
\usepackage{amssymb}
\usepackage{array}
\usepackage{bbding}
\usepackage{booktabs}
\usepackage{cite}
\usepackage{enumitem}
\usepackage{makecell}
\usepackage{mathptmx}
\usepackage{microtype}
\usepackage{multirow}
\usepackage{pifont}
\usepackage{placeins}
\usepackage{tabularx}
\usepackage{textcomp}
\usepackage{longtable}
\usepackage{float}
\usepackage{caption}
\usepackage{subcaption}
\usepackage{graphicx}
\usepackage{arydshln}
\usepackage{multicol}
\usepackage{pdfpages}
\usepackage{graphicx}
\usepackage[table,xcdraw]{xcolor}

\definecolor{PosColor}{HTML}{EF654E}
\definecolor{NeuColor}{HTML}{6877FF}
\definecolor{NegColor}{HTML}{39C6A5}
\DeclareCaptionStyle{ruled}{labelfont=normalfont,labelsep=colon,strut=off} 
\floatstyle{ruled}
\newfloat{listing}{tb}{lst}{}
\floatname{listing}{Listing}

\title{InfoAffect: Affective Annotations of Infographics in Information Spread}

\author{
  Zihang Fu\textsuperscript{\rm 1},
  Yunchao Wang\textsuperscript{\rm 1}\thanks{Corresponding author},
  Chenyu Huang\textsuperscript{\rm 1}\\
  Guodao Sun\textsuperscript{\rm 1, 2},
  Ronghua Liang, \textit{Senior Member, IEEE}\textsuperscript{\rm 1}
}

\affiliations{
  \textsuperscript{\rm 1}College of Computer Science and Technology, Zhejiang University of Technology\\
  \textsuperscript{\rm 2}Zhejiang Key Laboratory of Visual Information Intelligent Processing\\
  {\{fuzihang, guodao, rhliang\}@zjut.edu.cn},
  wyctears@gmail.com,
  hcy1776848600@gmail.com
}

\begin{document}

\maketitle

\begin{figure*}[t]
  \hspace*{-0.33cm}
  \centering
  \includegraphics[width=\textwidth]{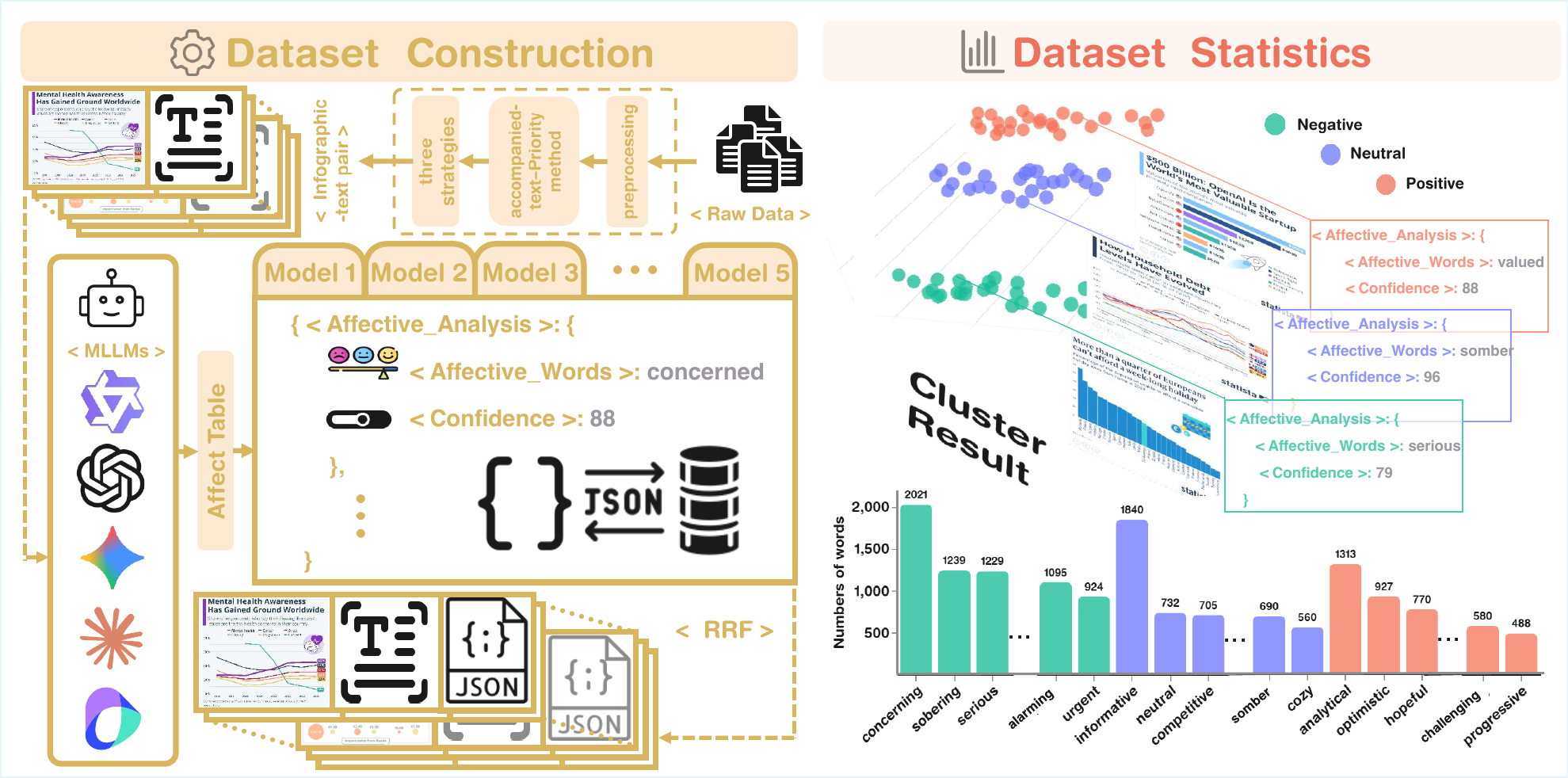}
  \caption{This provides an overview of the \textit{\textbf{InfoAffect}} dataset, illustrating its dual-modal structure that integrates visual infographics and their accompanied texts. It summarizes the overall workflow, including data collection, preprocessing, affect extraction via MLLMs, and fusion through Reciprocal Rank Fusion.}
  \label{fig:teaser}
\end{figure*}

\begin{abstract}
Infographics are widely used in social media to convey complex information, yet how they influence users' affects remains underexplored due to the scarcity of relevant datasets.
To address this gap, we introduce a 3.5k-sample affect‐annotated \textit{InfoAffect} dataset, which combines textual content with real-world infographics.
We first collected the raw data from six fields and aligned it via preprocessing, the accompanied-text-priority method, and three strategies to guarantee quality and compliance.
After that, we constructed an \textit{Affect Table} to constrain annotation.
We used five state-of-the-art multimodal large language models (MLLMs) to analyze both modalities, and their outputs were fused with Reciprocal Rank Fusion (RRF) algorithm to yield robust affects and confidences. 
We conducted a user study with two experiments to validate usability and assess \textit{InfoAffect} dataset using the Composite Affect Consistency Index (CACI), achieving an overall score of 0.608, which indicates high accuracy.The \textit{\textbf{InfoAffect}} dataset is available in a public repository at \url{https://github.com/bulichuchu/InfoAffect-dataset}.
\end{abstract}


\section{Introduction}
In contemporary social media contexts, information presentation is shifting from text-only formats toward complex multimodal narratives~\cite{amit2018digital}.
Infographics, which represent information, data, imagery or knowledge~\cite{lankow2012infographics,newsom_public_2008}, have emerged as a central medium through which social media users acquire complex information, participate in public discussions, and engage in political mobilization~\cite{chang_characterizing_2023}.
They effectively promote information sharing and engagement across diverse social media users~\cite{chang_characterizing_2023}.
It could present complex data in ways that are both intuitive and visually engaging~\cite{smiciklas2012power,lankow2012infographics}.
For instance, a recent study~\cite{lee_perceptions_2022} found that infographic posts on social media outperformed text-only articles in shareability and comprehension, with 90\% of users indicating enhanced understanding through visual presentations.

Although considerable progress has been made in the analysis of infographics~\cite{jacob2020visualising,kunze2021infographics}, research on the affective dimensions of infographics, such as people's emotions, moods, or feelings, remains scarce and underexplored~\cite{lan2021smile}.
One approach to advancing affective analysis in infographics is to develop datasets that capture the affects elicited by various infographic designs.
Previous work~\cite{lan2021smile} compiled a corpus via manual collection and crowdsourced annotations, laying a foundation for subsequent studies in this area.
However, the social media and visualization research communities currently lack affects annotated dual-modal infographic datasets that incorporate both visual content and textual information.
The absence of such datasets limits our ability to systematically study how visual design and textual description jointly shape users' affects in social media contexts.

\begin{table*}[htbp]
\definecolor{customgreen}{rgb}{0.117, 0.737, 0.239}
\definecolor{customred}{rgb}{0.85,0.13,0.13}
\definecolor{customfalse}{RGB}{212,101,96}
\definecolor{cellhi}{HTML}{FFE6E6}
\newcommand{\cmark}{{\textcolor{customgreen}{\ding{51}}}}
\newcommand{\xmark}{{\textcolor{customred}{\ding{55}}}}
\newcommand{\hi}[1]{\cellcolor{cellhi}{#1}}
\renewcommand\tabularxcolumn[1]{>{\centering\arraybackslash}m{#1}}
\caption{Overview of existing infographics datasets. 
}
\centering
\renewcommand{\arraystretch}{1.25}
\begin{tabularx}{\textwidth}{l
>{\centering\arraybackslash}X
>{\centering\arraybackslash}m{2.2cm}
>{\centering\arraybackslash}X
>{\centering\arraybackslash}X
>{\centering\arraybackslash}X
>{\centering\arraybackslash}X}

\Xhline{1.2pt}
\textbf{Dataset} & \textbf{Size} & \textbf{Affect Granularity} &\textbf{Affect Label} & \textbf{Text Modality}& \textbf{Real-world Image}  & \textbf{Structured Output} \\ \cline{1-7}
\textit{ChartGalaxy}~\cite{li2025chartgalaxy}      & 1.76M & \xmark & \xmark& \xmark & \xmark  & \cmark \\
\textit{InfoDet}~\cite{InfoDet}                    & 101K   & \xmark & \xmark& \xmark & \xmark & \xmark \\
\textit{InfographicVQA}~\cite{mathew2022infographicvqa} & 5.5K  & \xmark & \xmark& \xmark & \cmark  & \cmark \\
\textit{Kaggle Infographics Dataset}~\cite{Kaggle} & 7.88K & \xmark & \xmark& \xmark & \xmark  & \xmark \\
\textit{Visually29K}~\cite{madan_parsing_2021}       & 29K    & \xmark & \xmark& \cmark & \cmark  & \cmark \\
\textit{Smile or Scowl?}~\cite{lan2021smile}       & 0.9K    & \textcolor{customred}{Binary Affects} & \cmark& \xmark & \cmark  & \cmark \\
 \hline
\textit{InfoAffect} (\textbf{Ours})                               & 3.5K  & \textcolor{customgreen}{Ternary Affects} & \cmark& \cmark & \cmark  & \cmark \\ \Xhline{1.2pt}
\end{tabularx}
\label{tab:datasets}
\end{table*}

In response to this gap,  we introduce \textit{InfoAffect} dataset.
The design of \textit{InfoAffect} dataset offers three key advantages: First, we curate real-world infographics from six fields rather than synthetics, so that MLLMs trained or fine-tuned on \textit{InfoAffect} dataset learn from real-world design conventions, linguistic registers, and noise characteristics.
Second, we construct an \textit{Affect Table} to guide and constrain affect annotations, ensuring affects remain within a limited label space. This bounding reduces synonym drift and label proliferation.
Third, we combine accompanied texts with real-world infographics to mirror how infographics are encountered in practice; this dual-modal design captures the interplay between visual design and linguistic framing and yields more faithful affect predictions than image-only or text-only approaches.
which can be used to advance the study of affective analysis.
In summary, the main contributions are as follows:
 \begin{itemize}
 \item We introduce the \textit{InfoAffect} dataset, which comprises \textbf{3.5k} infographic–text pairs, each annotated with affect labels and corresponding confidence scores.
 
 \item We introduce a systematic method for extracting affects from infographic–text pairs, enabling the identification of specific affective expressions.
 
 \item We conduct a user study validating the usability and accuracy of the \textit{InfoAffect} dataset. The results show that it is highly useful for real-world infographic design and achieves high accuracy.
 \end{itemize}


\section{Related Work}
In this section, we review the previous works. Specifically focusing on three key areas: \textit{Theories of Affects}, \textit{Existing Infographics Datasets} and \textit{Affect Recognition Methodology}.

\subsection{Theories of Affects}
The classic affects discrete theory posits the existence of a finite set of affective categories that are universally applicable. Each category is associated with distinct features~\cite{ekman_argument_1992,noauthor_general_1980}.
Subsequently, dimensional accounts reconceptualize affect as coordinates in a small number of continuous axes, and affective categories treated as vague regions~\cite{mehrabian_pleasure-arousal-dominance_1996}.
Later, appraisal theories propose that affects arise from evaluative checks of events relative to an individual's goals, norms, and resources, yielding patterned yet context sensitive response profiles~\cite{scherer_what_2005}.
Over the past decade, constructionist accounts argue that affect categories are not fixed modules but contextually constructed interpretations of core affect using conceptual knowledge and language, predicting substantial intra-category variability~\cite{barrett2017emotions}.

However, recent studies have demonstrated limitations of discrete affect models in complex tasks such as music emotion recognition~\cite{eerola_comparison_2011}.
Most recently, Semantic Space Theory (SST) has offered a data-driven alternative to prespecified affect taxonomies~\cite{cowen2019mapping, cowen2020face, cowen2019speech}. 
Instead of assuming a fixed set of discrete categories, SST aggregates rich human language, including free descriptions, comparative judgments, and similarity ratings, to embed both felt and perceived affects in a high-dimensional semantic space. In this representation, affects are organized continuously: nearby points indicate fine-grained experiential similarity, and broader regions arise as coherent neighborhoods without being imposed in advance. The geometry captures smooth transitions along affective gradients, preserves contextual nuance through local topology, and admits overlapping groupings that better reflect human judgments. 
Given the multimodal nature of infographics and their capacity to evoke complex affects, we ground our approach in SST and use its continuous classification and analysis of affects.

\subsection{Affect Recognition Methodology}

Early approaches to affect recognition relied heavily on unimodal inputs, such as facial-expression classifiers mapping images to affect labels~\cite{dino2019facial} or audio-based sentiment detection~\cite{kaushik2017automatic}, typically evaluated in constrained settings with limited variability.
However, these early methods struggled to generalize across real-world, multimodal contexts~\cite{pan2023multimodal}, where affect is conveyed jointly through visuals, language, and sound and where distribution shift, occlusion, and noise are common.
Intermediate approaches explored different fusion timing and modalities, text, image and audio, with improved robustness in naturalistic stimuli~\cite{pan2023review}. 
More recent work therefore adopted fusion strategies that combine multiple modalities to enhance affect‐recognition performance~\cite{pan2023multimodal}, leveraging complementary cues and reducing reliance on any single modal.
The latest wave leverages deep learning architectures such as large multimodal pretrained models: they encode textual and visual inputs, apply cross-attention layers to align semantics between modalities, and output affect labels or continuous affect dimensions~\cite{ahnert_extracting_2025-1}.

In infographic affect analysis, the modality combination is distinctive: visual design and textual description co-exist within a single artifact and jointly shape the viewer's affect, yet prior methods seldom target this pairing directly for affect extraction tasks~\cite{khatiwada_towards_2025}. Therefore, we develop a unified pipeline tailored to this setting that combines insights from earlier unimodal and more recent multimodal advances. Rather than committing to a single modeling recipe, the pipeline synthesizes established practices with contemporary architectures under a common formulation. In spirit, it is comprehensive rather than prescriptive: it draws on what the literature has already shown to be effective, accommodates newer techniques as they mature, and provides a consistent basis for interpreting results in terms of both visual design and textual framing. This integration allows us to treat infographic affect as a genuinely multimodal phenomenon while retaining the flexibility needed for diverse tasks, datasets, and evaluation protocols.

\subsection{Existing Infographics Datasets}\label{sec:existingDataset}
Advances in infographic research have produced a variety of datasets that differ significantly in scale, modality, and supervision.
Datasets like \textit{ChartGalaxy} (1.76M) and \textit{InfoDet} (101K) focus on visual structure but lack affective labels and textual modalities, making them limited in their ability to capture affects in infographics~\cite{li2025chartgalaxy,InfoDet}.
\textit{InfographicVQA} (5.5K) provides structured question–answer pairs for real-world infographics, but it targets visual question answering rather than affective analysis and lacks both affect annotations and a dedicated text modality~\cite{mathew2022infographicvqa}.
The \textit{Kaggle Infographics Dataset} (7.88K), primarily synthetic or web-scraped, suffers from inconsistent labeling~\cite{Kaggle}, while ``\textit{Smile or Scowl?}'' (0.9K) offers binary affect labels but relies solely on the visual modality~\cite{lan2021smile}.

To the best of our knowledge, \textit{InfoAffect} dataset is the only affect-annotated dataset that combines textual content with real-world infographics; although its overall scale is moderate, it offers broad coverage across affective dimensions. This unique combination enables a deeper exploration of how visual design and textual cues interact to shape affects, addressing a critical gap in the current infographic research landscape. A side-by-side summary of these datasets is provided in Table~\ref{tab:datasets}.

\section{Dataset Construction}

In this section, we describe the full process of \textit{InfoAffect} dataset construction, from data preparation and affects extraction, through prompt based affect extraction using MLLMs, to formatting and fusion of the final affective outputs, as shown in the left panel of Figure~\ref{fig:teaser}.

\subsection{Data Preparation}\label{sec:data-prep}
We gathered a large and heterogeneous corpus of 10,000 infographics and their accompanied text from six distinct fields to ensure coverage, diversity, and ecological validity.
This topical diversity supports cross-domain evaluation and reduces overfitting to specific content genres, following best practices in dataset and transparency~\cite{gebru_datasheets_2021,hutchinson_towards_2021}.
All data collection was conducted in strict accordance with applicable authorization agreements, robots directives, and platform terms of use.
Specifically, the source fields were as follows:
\begin{itemize}
    \item \textbf{social media posts:} Infographics shared on social media platforms like X, such as an infographic illustrating the thermal wind balance~\cite{X_infographic}.
    \item \textbf{Data journalism:} Online news platforms that regularly publish data-driven reports, such as Wall Street Journal~\cite{wsj2015};
    \item \textbf{Governmental portals:} Official websites that provide publicly accessible reports, such as Smart city of HK~\cite{hk25};
    \item \textbf{Non-governmental organizations:} Research-driven institutions that publish reports, policy briefs, and white papers, such as ReliefWeb~\cite{ReliefWeb};
    \item \textbf{Academic and institutional repositories:} Digital archives and platforms that host scientific research and educational content, including infographics aimed at making complex research more accessible to the public, as discussed in Sect.~\textit{Existing Infographics Datasets};
    \item \textbf{Corporate and institutional reports:} Annual reports, sustainability reports, and financial statements from companies and organizations, typically released under open access or permissive terms, which include infographics to support business analysis and transparency, such as Adobe Digital Trends Report~\cite{adobe};
\end{itemize}

From each record, we retained the infographic and all accompanied texts, such as headline, standfirst, caption, or article snippet. 
After collecting the raw data, we performed several preprocessing steps,
all infographics were converted to a canonical RGB format and constrained not to exceed a specified maximum size.
We removed corrupted files and filter images below a minimum resolution threshold to avoid unreadable text. 
To control redundancy, we used perceptual hashing to detect near duplicates~\cite{hao_its_2021}. For each duplicate cluster, we kept the image with the highest total pixel count; if there was still a tie, we kept the earliest by timestamp.
When infographics were embedded in PDF or HTML, we extracted the focal figure region to support downstream alignment.

After preprocessing, an accompanied-text–priority method was applied to the remaining 3,500 infographics. This ensures each infographic was accompanied by faithful textual summaries. We mapped each infographic to a concise, content-faithful textual description that reflected its communicative intent.
\ding{182}
\textbf{Accompanied-text prioritization.} If accompanied text was available, we prioritized it as the primary source.
\ding{183}
\textbf{Prompted Description.} Otherwise, if accompanied text was absent (approximately 11\% of the dataset), we instead used a well-designed prompt to guide GPT-4o to produce the description. This text is concise, factual, and style-consistent. 
The complete prompt is provided in (Appendix~\ref{sec:Dataset_Preparation_Prompts}).

Finally, we implemented three strategies to guarantee the quality and compliance of the data and text. 
Each data instance consists of an infographic-text pair, where the image and accompanying text were treated as a unified entity. This structure provided a consistent multimodal representation, which served as the foundation for the subsequent affect analysis.
We performed a \textbf{consistency check} by comparing key facts in the generated text against the corresponding image and flagging any discrepancies for removal or regeneration. We also conducted a \textbf{style and length check} to verify adherence to journalistic conventions and length constraints, and excluded templated phrasing and speculative markers. Additionally, we carried out \textbf{human review} through manual assessments on stratified samples across sources and topics to validate factual accuracy and the absence of personally identifiable or sensitive information.

\subsection{Affect Table}\label{sec:emo_lable}
\begin{figure}[htbp]
    \setlength{\intextsep}{22pt}
    \includegraphics[width=\linewidth]{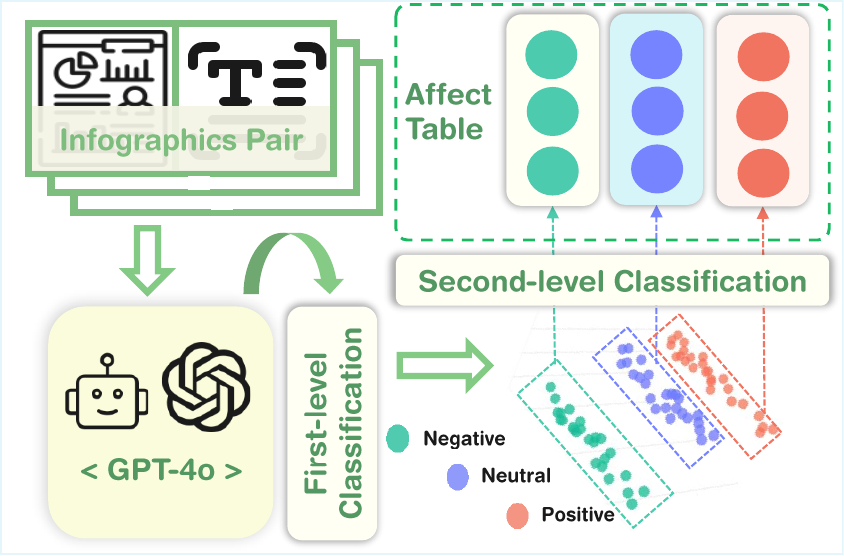}
    \caption{Overview of the \textit{Affect Table} construction process.}\label{fig:Table}
\end{figure}

To ensure that the affective output remains within a limited range, we designed an \textit{Affect Table} as a structured reference for affective extraction. The \textit{Affect Table} construction process is showed in Figure~\ref{fig:Table}.
We sampled 1,000 infographics from the previously introduced datasets~\cite{li2025chartgalaxy,InfoDet,mathew2022infographicvqa,Kaggle,lan2021smile}, and prompted GPT-4o~\cite{chatgpt4o2024} to extract affects, which were then organized through a \textit{two-level classification}. Based on this, we constructed the \textit{Affect Table}. The following paragraphs detail the \textit{two-level classification} process.

\paragraph{First-level Classification}
The \textit{first-level classification} aimed to provide an initial categorization of affects, establishing the coarse semantic structure for subsequent fine-grained clustering. 
While affective states in natural language are continuous and overlapping, most existing affective computing frameworks~\cite{mohammad_word_2022, buechel_word_2018, cowen2019mapping} converge on three dominant affective polarities—\textbf{positive}, \textbf{neutral}, and \textbf{negative}—as fundamental axes of affective evaluation. 
Following these established frameworks, the affects can be readily divided into three primary categories. However, such a coarse division was insufficient to capture the affective diversity. Therefore, we conducted a \textit{second-level classification} within each primary category to achieve a more fine-grained representation.

\paragraph{Second-level Classification}
In this phase, this study drew on the SST of affect~\cite{cowen2019mapping, cowen2020selfreport, cowen2021sst}, which posits that affects are continuously distributed in a high-dimensional space. Building on this theory, we projected each word into a point in a high-dimensional  affective space using Word2Vec~\cite{mikolov_efficient_2013}. 
To identify affective categories from these points, we employed Hierarchical Density-Based Spatial Clustering of Applications with Noise (HDBSCAN)~\cite{mcinnes_hdbscan_2017}, an unsupervised density-based clustering algorithm that did not require pre-specifying the number of clusters. 
Given a set of affective points \( \mathbf{A} = \{\mathbf{A}_1, \mathbf{A}_2, \dots, \mathbf{A}_n\} \subset \mathbb{R}^{m} \), where 
\(m\) was the embedding dimension, the algorithm groups pointed into clusters of varying densities, while labeling ambiguous or isolated words as noise.
For each affective point \( \mathbf{A}_i \), we computed its \textit{core distance} as
\begin{equation}
\mathrm{core}_k(\mathbf{A}_i) = \operatorname{dist}\big(\mathbf{A}_i, \mathbf{\textit{k}\text{NN}(i, k)}\big)
\end{equation}
where \(\mathbf{\textit{k}\text{NN}(i, k)}\) denotes the $\textit{k}$-Nearest Neighbors of point \( \mathbf{A}_i \) and \( \operatorname{dist}(\cdot) \) is measured by the Euclidean distance.
The algorithm then defines the \textit{mutual reachability distance} between two points \( \mathbf{A}_i \) and \( \mathbf{A}_j \) as
\begin{equation}
d_{\mathrm{mreach}}(\mathbf{A}_i, \mathbf{A}_j) = \max \big( \mathrm{core}_k(\mathbf{A}_i), \, \mathrm{core}_k(\mathbf{A}_j), \, \operatorname{dist}(\mathbf{A}_i, \mathbf{A}_j) \big)
\end{equation}
Based on these pairwise distances, a weighted graph is constructed, where each word is a node and the edges are weighted by \( d_{\mathrm{mreach}}(\mathbf{A}_i, \mathbf{A}_j) \). A minimum spanning tree (MST) is then computed over this graph, representing the most compact connections between affects. By progressively removing edges from the MST in order of decreasing density (increasing distance), a hierarchical clustering structure, known as the \textit{condensed cluster tree}, is formed. Each branch in this tree represents a potential cluster at a specific density level.
For each cluster \( C \), HDBSCAN measures its \textit{stability} as
\begin{equation}
\mathrm{Stability}(C) = \sum_{i \in C} \big( \lambda_{\text{death}}(i) - \lambda_{\text{birth}}(i) \big)_{+}
\end{equation}
where \( \lambda_{\text{death}}(i) = 1 / d_{\text{death}}(i) \) and \( \lambda_{\text{birth}}(i) = 1 / d_{\text{birth}}(i) \).
Here, \( d_{\text{death}}(i) \) and \( d_{\text{birth}}(i) \) are the mutual reachability distances at which point \( i \) joins and leaves cluster \( C \) along the cluster tree, respectively. HDBSCAN automatically retains clusters with higher stability as meaningful affective groups, while less stable ones are pruned as noise.
For each resulting cluster \( C_k \), we compute its centroid as the average of all word vectors in that cluster:
\begin{equation}
\mathbf{a}_k = \frac{1}{n_k} \sum_{\mathbf\mathbf{A}_i \in C_k} \mathbf\mathbf{A}_i
\end{equation}
where \( n_k = |C_k| \) is the number of words in cluster \( k \).
The centroid \( \mathbf{a}_k \) serves as the \textit{affective label}, representing the overall affective meaning of cluster \( C_k \).

\begin{figure}[t]
    \includegraphics[width=\linewidth]{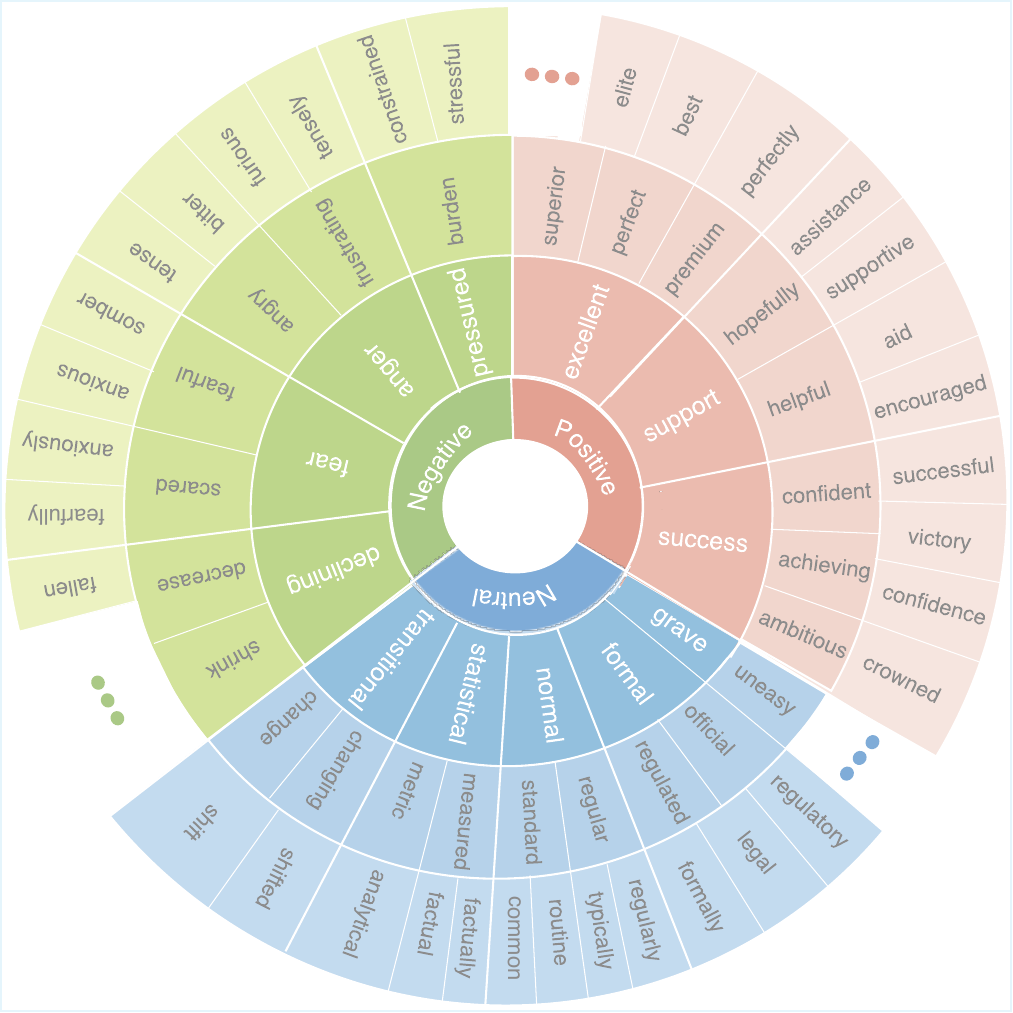}
    \caption{The innermost layer represents the three primary affective polarities classified by \textit{first-level affective classification}, while the outer rings display the finer \textit{second-level affective classification} that form the \textit{Affect Table}. Each sector's area corresponds to the relative frequency of affects, revealing a balanced and comprehensive affective coverage across different polarities.}
    \label{fig:sunburst}
\end{figure}

\textit{InfoAffect} dataset exhibited balanced coverage across the three affective polarities and a diverse set of subcategories derived from the SST of affect. As shown in Figure~\ref{fig:sunburst}, the hierarchical organization of categories was visualized via a sunburst chart, revealing both broad coverage and structured interrelationships. The balanced radial distribution indicated that the dataset was not overly biased toward any single polarity, while the layered structure supported the internal consistency of the \textit{two-level classification} and showed that the \textit{Affect Table} preserves categorical distinctions alongside semantic continuity.\label{sec:additional-characteristics}
The overall classification results were presented in (Appendix~\ref{sec:affect-table}) and Figure~\ref{fig:sunburst}; only the \textit{second-level classification} formed the final \textit{Affect Table}. A detailed account of how the model leveraged this table for supervised affect extraction was provided in Sect.~\textit{Prompting Strategies for Affect Extraction}.

\subsection{Affect Extraction with MLLMs}\label{sec:affect-extraction}
To perform supervised affect extraction based on the constructed \textit{Affect Table}, we employed 5 state-of-the-art MLLMs: GPT-4o~\cite{chatgpt4o2024}, Claude Sonnet 4~\cite{Claude4}, Doubao-1.5-Pro~\cite{doubao}, Gemini 2.5~\cite{gemini}, and Qwen3-VL-Plus~\cite{Qwen3-VL}. These models serve as independent annotators that analyze each infographic–text pair and extract affects guided by the \textit{Affect Table}. Prior research has validated that multi MLLMs annotations can achieve reliability comparable to ground truth~\cite{khatiwada_towards_2025}. Their outputs were subsequently fused using the RRF algorithm, as described in the following sections.

\subsubsection{JSON Schema for Affects}
The JSON schema used in \textit{InfoAffect} dataset follows a structured key–value schema to represent the affective analysis results of each infographic and its corresponding textual description. (Appendix~\ref{sec:JSON}) described the function of each key-value pair in the JSON file. For every infographic–text pair, we provided a single JSON file containing affect-related annotations under the ``\texttt{affective\_analysis}'' key. Each file included a list of affects and their corresponding confidence scores within the ``\texttt{affective\_words}'' array. This design enabled researchers to easily access both the lexical and quantitative dimensions of affective representation. 

\subsubsection{Prompting Strategies for Affect Extraction}\label{sec:prompting}

Prompt frameworks have demonstrated their effectiveness in optimizing interactions with MLLMs, thereby achieve more faithful instruction following~\cite{co-star2024,wang_taleframe_2025}. 
In this study, we employed the TIDD-EC prompt framework~\cite{tidd-ec2024}, which was composed of five essential components: \textit{task type}, \textit{detailed instructions}, \textit{does and don'ts}, \textit{examples}, and \textit{user content}.
By systematically defining the task scope, providing explicit operational guidelines, specifying behavioral constraints, illustrating desired results through examples, and incorporating user specific contextual information, the TIDD-EC prompt framework contributed to generating responses that are both accurate and relevant.
We maintain MLLMs were instructed to select affects exclusively from the \textit{Affect Table} introduced in Sect.~\textit{Affect Table}, rather than producing words freely. Each selected affect was required to be accompanied by a corresponding confidence score.
(Appendix~\ref{sec:t3}) summarizes the detailed prompt configurations applied to the affect extraction task under the TIDD-EC prompt framework.

\subsubsection{Affective Result Fusion}

\begin{figure}[!htbp]
    \includegraphics[width=\linewidth]{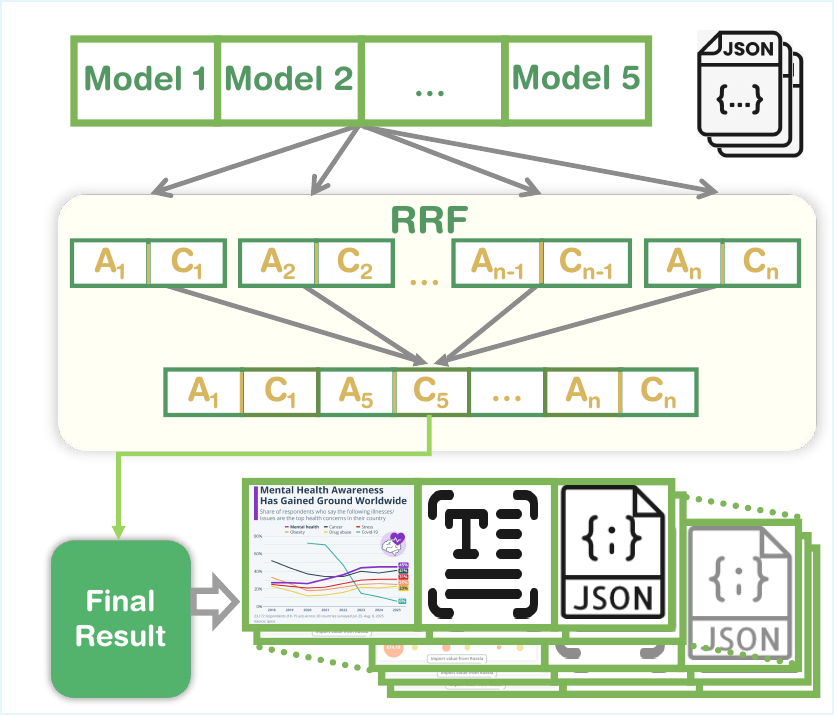}
    \caption{Detailed illustration of the Reciprocal Rank Fusion (RRF) algorithm applied in affective result fusion.}
    \label{fig:RRF}
\end{figure}

To mitigate potential biases inherent in individual MLLM and fully leverage the complementary strengths of different models, we integrated the results from the five models introduced in the beginning of Sect.~\textit{Affect Table} using the RRF algorithm~\cite{clarke2008novelty}. 
RRF algorithm is an effective fusion technique that requires no training and does not depend on initial scores.
This fusion strategy has been widely demonstrated to perform robustly across multimodal and retrieval tasks, effectively balancing precision and recall while enhancing the stability and consistency of final affective predictions~\cite{samuel2025mmmorrf}.
The fusion process was explained in Figure~\ref{fig:RRF}.

Supposed the output lists of five models are denoted as $M_1, M_2, M_3, M_4,$ and $M_5$.
For each affect $A$, its ranking position within model $i$ is represented as $r_{i(A)}$, and the associated confidence score assigned by that model was denoted as $C_{i(A)}$.
To consolidate the judgments of five models, a fused score for each affect was computed as follows:
\begin{equation}
\text{s}(A) = \sum_{i=1}^{5} \frac{C_{i(A)}}{k + r_{i(A)}}
\end{equation}
The constant $k$ serves as a smoothing factor that limits the influence of lower-ranked results on the final score, thereby reducing noise caused by less relevant predictions. In this work, $k$ was fixed to 0.
If a particular affect $A$ was absent from the output of model $i$, its rank $r_{i(A)}$ was regarded as zero, hence the $\text{s}(a)$ became infinite, implying that this model did not contribute to the computation of the fused score for that affect.
Finally, all affective terms were ranked in descending order according to their $\text{s}(A)$ values. 
This result reflected a comprehensive consensus that balanced and complemented the affective predictions of the five models.

\section{User Study}
\begin{figure*}[h]
    \centering
    \includegraphics[width=0.95\textwidth]{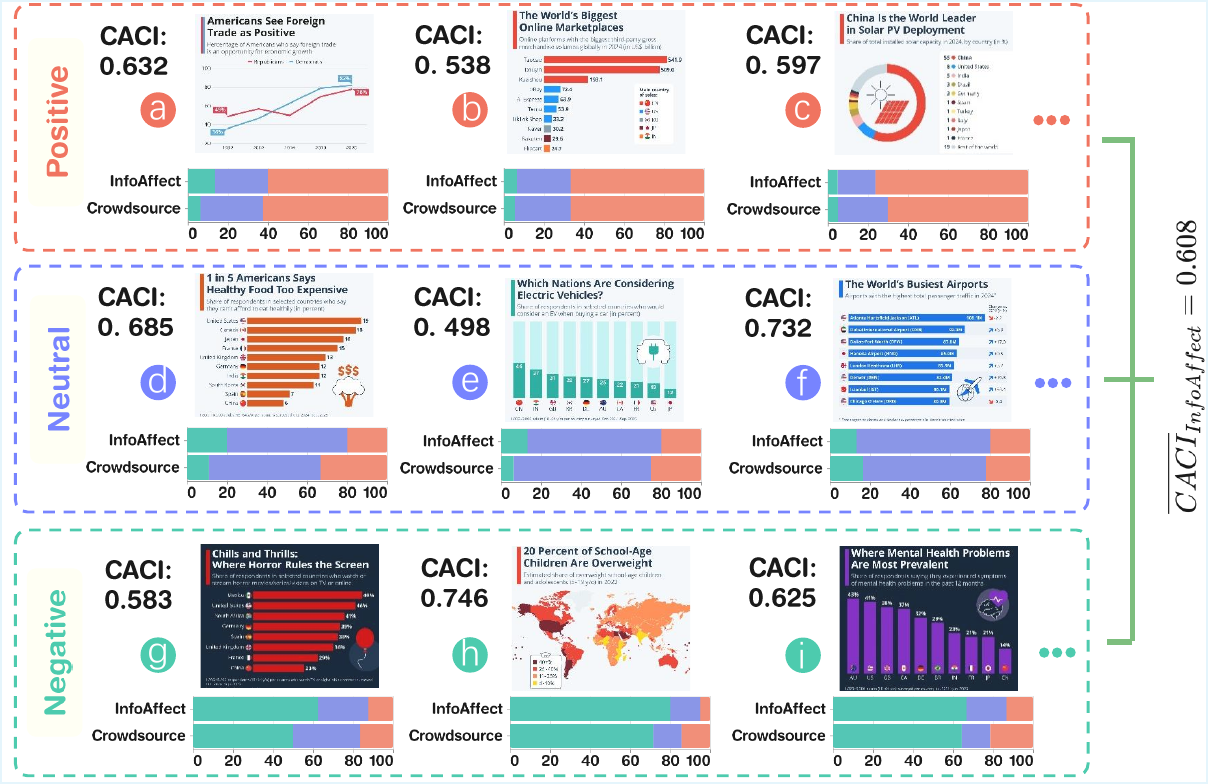}
    \caption{Visual comparison of CACI results between \textit{InfoAffect} dataset and crowdsourced ratings. 
Panels \textbf{(a–i)} were the representative examples cited in this section (labels on each panel); they illustrate typical cases rather than the full dataset. 
Each row corresponds to one affective category, indicated by colored legend markers: 
\textcolor{NegColor}{$\blacksquare$}~\textit{Negative}, 
\textcolor{NeuColor}{$\blacksquare$}~\textit{Neutral}, and 
\textcolor{PosColor}{$\blacksquare$}~\textit{Positive}. 
The bar charts below each panel represent the proportion of affective words from each affect category.
}
    \label{fig:user_study}
\end{figure*}

In this section, we elaborated on the detailed design, implementation process, participant composition, data collection methods, and result analysis of the practical case study, aiming to verify the usability and accuracy of the \textit{InfoAffect} dataset. Representative examples and the corresponding analytics for this section were illustrated in Figure~\ref{fig:user_study}.

\subsection{Participants}
We conducted a user study with 12 participants who had expertise in data journalism and visualization. Participants were recruited through Zhejiang University of Technology and local industry partners, which comprised two university faculty members (E1–E2) who teach courses on data visualization, three students (P1–P3), and seven industry practitioners (P4–P10), including data analysts, open-data officers, policy researchers, data consultants, marketing analyst, independent designers and design blogger. (Appendix~\ref{sec:participants}) summarizes the participant demographics, including gender, typical usage scenarios, and age. Participants ranged from 26 to 33 years old (mean = 29.3) and included five females and seven males. This user study received approval from the Institutional Review Board (IRB) of our university, all participants provided informed consent, and each received \$15 compensation.

\subsection{Exp.~\uppercase\expandafter{\romannumeral 1}
: Evaluating Usability of \textit{InfoAffect} dataset}
This section introduced the Exp.I, which investigates how the \textit{InfoAffect} dataset can assist creators in validating and correcting their intended affects within infographic design.

\subsubsection{Study Procedure}
In this study, each participant was invited to bring several of their previously created infographic for analysis.  
Before the experiment, participants were asked to recall and specify the affective original intent they had during the creation process, whether they aimed to express a particular affect or to maintain an affectively neutral tone.  This original intent was recorded for subsequent analysis.

Following the same processing pipeline used in constructing the \textit{InfoAffect} dataset described in Sect.~\textit{Affect Extraction with MLLMs}, we analyzed each infographic to extract its affects. The pipeline automatically extracted affects, which were then compared with the participants' original intent.
After that, We presented the extracted affects to the participants and conducted a semi-structured interview. The specific questions and answers were provided in (Appendix~\ref{sec:interview}).
Interviews were conducted with participants in a quiet setting, audio-recorded with informed consent, and supplemented by contemporaneous field notes capturing key quotations, clarifications, and contextual observations. Our analysis focused on how participants interpreted the detected affects, the reasoning they offered for those interpretations.

\subsubsection{Intent \emph{vs.} Observation}
We contrasted each participant’s original intent for their infographic with the affect extracted by the \textit{InfoAffect} dataset construction pipeline. We then discussed these contrasts in semi-structured interviews. Representative examples are shown in Figure~\ref{fig:user_study}. The contrasts fall into two patterns:

\paragraph{Validating}
The validation aspect examines whether the original intent of infographic creators is accurately realized in their visual designs. 
For those who aimed to convey a particular affect, we validate whether their original intent are effectively communicated. 
In contrast, for creators who intend to maintain an affectively neutral, this aspect assesses whether their works remain truly neutral without introducing unintended affects. 

\newcommand{\circled}[1]{\raisebox{.5pt}{\textcircled{\raisebox{-.5pt}{#1}}}}
For P6, she offered \circled{a} and reported aiming for ``\textit{hopeful}'', seeking a positive affect. The extracted affect is \textit{optimistic}. In the interview she noted, ``\textit{My intent in the policy report was to highlight a hopeful trend for policymakers' consideration, and the extracted affect of the infographic was `optimistic', which closely aligned with my original intent.}''
For P4, he offered \circled{e} and aimed to ``\textit{naturally present the information}'', simply enumerate which countries are more open to electric vehicles without taking an evaluative stance. The extracted affect was a neutral label \textit{informational}. He commented, ``\textit{My goal was just to report which nations are considering electric vehicles; the analysis produced neutral terms that align with the report-style tone I intended.}''

\paragraph{Correcting}
The correction aspect focuses on situations where discrepancies arise between the creator's intended affect and the extracted affects. 
Instead of directly modifying the design, the dataset provides an affective reference that helps creators recognize potential inconsistencies between their intended and extracted affects. 
Participants are then encouraged to adjust their own works—either by reinforcing insufficient affects or by reducing unintended ones. 

For P3, she offered \circled{h} and explained that her aim was to ``\textit{list the relevant figures in my manuscript}'' keeping the tone descriptive rather than evaluative. However, affects extracted by the \textit{InfoAffect} dataset were negative terms: \textit{alarming}, \textit{concerning}. In the interview she reflected, ``\textit{I only meant to report the numbers, but the result reads as negative; I suspect it's due to the warm red–orange ramp and the strong contrast in the legend.}''
For P10, he offered \circled{b} and said his intent was ``\textit{to express a negative winner-takes-all trend, namely a worrying concentration in which one platform dominates the others.}'' By contrast, the extracted affects were positive terms: \textit{confident}, \textit{competitive}. In the interview he reflected, ``\textit{I suspect the bright, saturated palette and clean ranking layout make it feel like achievement rather than imbalance; the long red top bar reads as `leading' more than `concerning' and the crisp flags and labels add a celebratory tone.}''

Across both aspects, the \textit{InfoAffect} dataset proved practically useful by contrasting participants' original affect with the extracted affect. In validating cases, it offered assurance and a defensible record that the original affect, whether positive or deliberately neutral, had been achieved (P7: ``\textit{reassuring to have an external read that matches what I meant}''). In correcting cases, it catalyzed focused reflection and concrete iteration, participants reported adjusting palettes and contrast, widening neutral ranges, simplifying labels, and tightening grids, so that the visual rhetoric realigns with the original intent. In short, the analysis confirms the usability of the \textit{InfoAffect} dataset.

\subsection{Exp.~\uppercase\expandafter{\romannumeral 2}
: Evaluating accuracy of \textbf{\textit{InfoAffect}} dataset}

This section introduced the Exp.II, which aimed to quantitatively assess the accuracy of affective extraction performed by \textit{InfoAffect} dataset.  
The goal was to verify whether the \textit{InfoAffect} results were consistent with human perceptions when evaluated by a general audience.

\subsubsection{Study Procedure}
Two sources of infographics were selected for this study:
\ding{182} 430 items randomly sampled from the constructed \textit{InfoAffect} dataset. \ding{183} 10 infographics collected in Exp~\uppercase\expandafter{\romannumeral 1}, in which discrepancies were observed between the participants' intended affects and the results of \textit{InfoAffect}. In total, 440 infographics were included in the study.
We used Amazon Mechanical Turk (MTurk) to gather responses from a large and diverse pool of participants.
32 participants were permitted to complete this task after a 5-minute standardized introduction. We paid each participant \$1.5 for labeling 10 infographics. Additionally, we included the 12 participants from Exp~\uppercase\expandafter{\romannumeral 1}, yielding a total sample of 44 unique participants.
Each infographic was presented to participants for affective evaluation.
They were asked to annotate affects and confidence for each infographic.

\subsubsection{CACI Analysis}
 To evaluate the agreement between \textit{InfoAffect} results and human perceptions, we proposed a Composite Affect Consistency Index (CACI) that combines the dominant affect agreement and precision of top-ranked affects.For each infographic $i$, we first compute two indicators: the dominant class match $M_i$ and the retrieval success $S_i(k)$ at top-$k$. The overall CACI score is then obtained by averaging the product of these two indicators across all infographics.

Formally, CACI is defined as:

\begin{equation}
\label{eq:CACI}
\text{CACI}(k) = \frac{1}{N}\sum_{i=1}^{N} M_i \times S_i(k)
\end{equation}

Where $N = 440$ is the total number of infographics mentioned above, and we evaluate CACI at $k \in \{1, 3, 5\}$.We define the dominant class match indicator $M_i$ based on three-dimensional affect vectors. For each infographic $i$, let $(n_1^{(i)}, ne_1^{(i)}, p_1^{(i)})$ represent the number of raters selecting Negative, Neutral, and Positive in the \textit{InfoAffect} prediction, and $(n_2^{(i)}, ne_2^{(i)}, p_2^{(i)})$ for human annotations. Then:

\begin{equation}
\label{eq:M}
M_i = 
\begin{cases}
1, & \text{if } \arg\max(n_1^{(i)}, ne_1^{(i)}, p_1^{(i)}) = \arg\max(n_2^{(i)}, ne_2^{(i)}, p_2^{(i)})\\
0, & \text{otherwise}
\end{cases}
\end{equation}

The retrieval success indicator $S_i(k) \in \{0, 1\}$ represents whether the top-ranked affect from human annotations appears in the top-$k$ predictions from \textit{InfoAffect}. Let $a_i^*$ denote the affect word with the highest confidence from the crowdsource results for infographic $i$, and let $\text{Top}_k^{(i)}$ denote the set of top-$k$ predictions from \textit{InfoAffect} for infographic $i$. Then $S_i(k)$ is computed as:

\begin{equation}
\label{eq:S}
S_i(k) = 
\begin{cases}
1, & \text{if } a_i^* \in \text{Top}_k^{(i)}\\
0, & \text{otherwise}
\end{cases}
\end{equation}

CACI effectively captures both the alignment of dominant affects and the precision of top-ranked affects, providing a comprehensive measure of affective consistency between \textit{InfoAffect} results and human perceptions. The multiplicative formulation in Equation~\ref{eq:CACI} ensures that both aspects must be satisfied for a high score: the dominant class match indicator $M_i$ acts as a binary gate that enforces overall affective alignment, while the retrieval success indicator $S_i(k)$ measures the accuracy of affect extraction. Notably, when the predicted dominant affect class (Positive, Neutral, or Negative) does not match human perceptions, $M_i$ becomes 0, which drives the contribution of that sample to 0 regardless of the value of $S_i(k)$. Conversely, when $M_i = 1$, the sample's contribution is determined solely by $S_i(k)$.

As shown in Figure~\ref{fig:user_study}, we present infographic samples together with their corresponding CACI values. Table~\ref{tab:caci} presents the CACI results for with and without the \textit{Affect Table} at different top-$k$. Each row shows $\text{CACI}(k)$ computed by averaging $M_i \times S_i(k)$ across all 440 infographics. The results show that the resulting $\text{CACI}_{\textit{InfoAffect}} = 0.608$ (averaged across $k \in \{1, 3, 5\}$), and incorporating the \textit{Affect Table} improved affect extraction by 13.4\% on average compared to the condition without the \textit{Affect Table}. These results collectively demonstrate that the affective extraction maintains high accuracy in identifying affects of infographics.

\begin{table}[t]
\centering
\caption{CACI results for \textit{InfoAffect} at different top-$k$. Each row shows $\mathrm{CACI}(k)$ computed by averaging $M_i \times S_i(k)$ across all 440 infographics.}
\label{tab:caci}
\renewcommand{\arraystretch}{1.25}

\begin{tabular}{|l|c|c|}
\hline
\multicolumn{1}{|c|}{\textbf{Evaluation}} &
\multicolumn{1}{c|}{\makecell{\textbf{Without}\\\textit{Affect Table}}} &
\multicolumn{1}{c|}{\makecell{\textbf{With}\\\textit{Affect Table}}} \\
\hline
$\mathrm{CACI}(1)$ & 0.372 & 0.454 {\small (+22.0\%)} \\
\hline
$\mathrm{CACI}(3)$ & 0.573 & 0.643 {\small (+12.2\%)} \\
\hline
$\mathrm{CACI}(5)$ & 0.664 & 0.726 {\small (+9.3\%)} \\
\hline
\textbf{Average}   & 0.536 & 0.608 {\small (+13.4\%)} \\
\hline
\end{tabular}
\end{table}

\section{Limitations and Future Work}
While \textit{InfoAffect} dataset advanced affective analysis of infographics, we acknowledged two important limitations. First, our dataset size remains modest. Second, while the \textit{first-level classification} results are balanced, the \textit{second-level classification} results are long-tailed due to the fine-grained classification and the use of real-world infographics.
In future work, we will study how specific visual and textual elements in infographics shape perceived affect by varying one factor at a time on existing designs and noting the affect shifts. 


\section{Conclusion}
We presented \textit{InfoAffect} dataset, a 3.5k-sample dual-modal that pairs real-world infographics with accompanied text and constrains affect labels with the \textit{Affect Table}.
The corpus covers six fields and is prepared with three methods to ensure alignment and quality.
We extract affects with five state-of-the-art MLLMs and fuse their outputs using RRF to obtain robust affective results.
We validated the resource with two experiments that assessed usability and accuracy of \textit{InfoAffect} dataset.
The accuracy study introduced the CACI and achieved an overall score of 0.608.
\textit{InfoAffect} dataset establishes a foundation for future research on affective understanding, model training, and affectively informed visualization design.

\section{Acknowledgments}
This work was supported in part by National Natural Science Foundation of China (62422607, 62372411, 62036009, 62432014), and Zhejiang Provincial Natural Science Foundation of China (LR23F020003).

\bibliography{aaai25}


\newpage
\section{Paper Checklist}

\newcommand{\answerYes}[1]{\textcolor{blue}{#1}} 
\newcommand{\answerNo}[1]{\textcolor{teal}{#1}} 
\newcommand{\answerNA}[1]{\textcolor{gray}{#1}} 
\newcommand{\answerTODO}[1]{\textcolor{red}{#1}} 
\begin{enumerate}

\item For most authors...
\begin{enumerate}
    \item  Would answering this research question advance science without violating social contracts, such as violating privacy norms, perpetuating unfair profiling, exacerbating the socio-economic divide, or implying disrespect to societies or cultures?
    \answerYes{Yes}
  \item Do your main claims in the abstract and introduction accurately reflect the paper's contributions and scope?
    \answerYes{Yes}
   \item Do you clarify how the proposed methodological approach is appropriate for the claims made? 
    \answerYes{Yes}
   \item Do you clarify what are possible artifacts in the data used, given population-specific distributions?
    \answerNA{NA (no population-specific human distributions are modeled in this work).}
  \item Did you describe the limitations of your work?
    \answerYes{Yes}
  \item Did you discuss any potential negative societal impacts of your work?
    \answerNo{No, because we haven’t found any potential negative societal impacts.}
      \item Did you discuss any potential misuse of your work?
    \answerNo{No}
    \item Did you describe steps taken to prevent or mitigate potential negative outcomes of the research, such as data and model documentation, data anonymization, responsible release, access control, and the reproducibility of findings?
    \answerYes{Yes}
  \item Have you read the ethics review guidelines and ensured that your paper conforms to them?
    \answerYes{Yes}
\end{enumerate}

\item Additionally, if your study involves hypotheses testing...
\begin{enumerate}
  \item Did you clearly state the assumptions underlying all theoretical results?
    \answerNA{NA (no hypothesis-testing theoretical results).}
  \item Have you provided justifications for all theoretical results?
    \answerNA{NA}
  \item Did you discuss competing hypotheses or theories that might challenge or complement your theoretical results?
    \answerNA{NA}
  \item Have you considered alternative mechanisms or explanations that might account for the same outcomes observed in your study?
    \answerNA{NA}
  \item Did you address potential biases or limitations in your theoretical framework?
    \answerNA{NA}
  \item Have you related your theoretical results to the existing literature in social science?
    \answerNA{NA}
  \item Did you discuss the implications of your theoretical results for policy, practice, or further research in the social science domain?
    \answerNA{NA}
\end{enumerate}

\item Additionally, if you are including theoretical proofs...
\begin{enumerate}
  \item Did you state the full set of assumptions of all theoretical results?
    \answerNA{NA (no theoretical proofs).}
	\item Did you include complete proofs of all theoretical results?
    \answerNA{NA}
\end{enumerate}

\item Additionally, if you ran machine learning experiments...
\begin{enumerate}
  \item Did you include the code, data, and instructions needed to reproduce the main experimental results (either in the supplemental material or as a URL)?
    \answerYes{Yes, dataset is released via a public repository.}
  \item Did you specify all the training details (e.g., data splits, hyperparameters, how they were chosen)?
     \answerNA{NA (no model training;)}
     \item Did you report error bars (e.g., with respect to the random seed after running experiments multiple times)?
    \answerNA{NA (no stochastic training runs / seed-based reporting described).}
	\item Did you include the total amount of compute and the type of resources used (e.g., type of GPUs, internal cluster, or cloud provider)?
    \answerNA{NA (no model training).}
     \item Do you justify how the proposed evaluation is sufficient and appropriate to the claims made? 
    \answerYes{Yes}
     \item Do you discuss what is ``the cost'' of misclassification and fault (in)tolerance?
     \answerNo{No}
  
\end{enumerate}

\item Additionally, if you are using existing assets (e.g., code, data, models) or curating/releasing new assets, \textbf{without compromising anonymity}...
\begin{enumerate}
  \item If your work uses existing assets, did you cite the creators?
      \answerYes{Yes (prior datasets/models/sources are cited).}
  \item Did you mention the license of the assets?
    \answerNo{No (licenses are not explicitly stated).}
  \item Did you include any new assets in the supplemental material or as a URL?
     \answerYes{Yes (dataset released via a public repository URL).}
  \item Did you discuss whether and how consent was obtained from people whose data you're using/curating?
    \answerNA{NA (data are collected from public sources).}
  \item Did you discuss whether the data you are using/curating contains personally identifiable information or offensive content?
    \answerYes{Yes}
\item If you are curating or releasing new datasets, did you discuss how you intend to make your datasets FAIR (see \citet{fair})?
    \answerYes{Yes}
\item If you are curating or releasing new datasets, did you create a Datasheet for the Dataset (see \citet{gebru2021datasheets})? 
\answerYes{Yes}
\end{enumerate}

\item Additionally, if you used crowdsourcing or conducted research with human subjects, \textbf{without compromising anonymity}...
\begin{enumerate}
  \item Did you include the full text of instructions given to participants and screenshots?
     \answerNo{No (only summarized in appendix).}
  \item Did you describe any potential participant risks, with mentions of Institutional Review Board (IRB) approvals?
    \answerYes{Yes (IRB approval and informed consent are stated).}
  \item Did you include the estimated hourly wage paid to participants and the total amount spent on participant compensation?
     \answerYes{Yes}
   \item Did you discuss how data is stored, shared, and deidentified?
   \answerYes{Yes}
\end{enumerate}

\end{enumerate}

\clearpage
\onecolumn
\appendix
\setcounter{secnumdepth}{2}

\section{Dataset Preparation Prompts}\label{sec:Dataset_Preparation_Prompts}
Task:
You are given an infographic image. Your task is to produce a concise, factual, and content-faithful textual description of the infographic.

Guidelines:

Focus strictly on the information conveyed by the infographic, including the main topic, key variables, data relationships, trends, comparisons, or notable patterns.

Describe what the infographic communicates, not how it looks. Avoid stylistic, aesthetic, or subjective language (e.g., colors, layout beauty, visual appeal).

Do not speculate beyond the visible content. Do not infer causes, motivations, or implications unless they are explicitly stated in the infographic.

If numerical values, time ranges, categories, or entities are present, summarize them accurately at a high level without exhaustively listing all values.

Write in a neutral, encyclopedic tone, similar to a figure caption or news standfirst.

The description should be self-contained, understandable without seeing the image, and no longer than 3–5 sentences.

Output:
A single paragraph describing the infographic’s communicative intent and key information.

\newpage
\section{Affect Table}\label{sec:affect-table}
Results of the \textit{two-level classification}. The left column corresponds to the \textit{first-level classification} results, the middle column contains the \textit{second-level classification} results, and the right column lists the affects initially generated by GPT-4o.

\newcolumntype{L}[1]{>{\raggedright\arraybackslash}m{#1}}
\newcolumntype{C}[1]{>{\centering\arraybackslash}m{#1}}
\newlength{\ColA}\setlength{\ColA}{18mm} 
\newlength{\ColB}\setlength{\ColB}{20mm} 
\newlength{\ColC}\setlength{\ColC}{\dimexpr\linewidth-\ColA-\ColB\relax}

\begin{table}[H]
\centering
\scriptsize
\setlength{\tabcolsep}{4pt}

\begin{tabular}{@{}C{\ColA}@{} L{\ColB}@{} L{\ColC}@{}}
\toprule
\textbf{First-level} & \textbf{Second-level} & \textbf{Affects} \\
\midrule
\multirow[c]{16}{*}{\rotatebox[origin=c]{90}{\textbf{Positive}}}
 & rise       & growth, grew, growing, expansion, expanding, rise, risen, rising, climb, ascending, expansive, expanded \\
 & soaring    & soared, soaring, surge, surged, surging, skyrocketed, booming, boost, explosive, buoyant, skyrocket, boosted \\
 & success    & success, successful, achievement, achieving, accomplished, triumph, victory, milestone, crowned, triumphant, victorious, dominant, dominance, confident, confidence, ambitious, ambition, competitive, competitiveness \\
 & excellent  & excellent, superior, exceptional, outstanding, top, highest, best, elite, premium, perfect, impressive, impressively, perfectly \\
 & innovative & innovative, groundbreaking, pioneering, trailblazer, revolutionary, visionary, modern, futuristic, progressive, progressively, modernized, transformative, transformatively \\
 & effective  & effective, efficient, productive, practical, useful, valuable, functional, proactive, resourceful, informative, informatively, professional, professionally, strategic, strategically, practically \\
 & reliable   & trust, trusted, trustworthy, reliable, authentic, stable, stability, secure, security, credible, credibility, authenticity, resilient, resiliently, significant, significantly \\
 & support    & support, supportive, helpful, aid, assistance, collaboration, cooperative, partnership, united, encouraging, encouragingly \\
 & joy        & happy, happiness, joyful, jubilant, cheerful, elated, exuberant, delightful, pleasant, sweet, joy, optimistic, optimistically, hopeful, hopefully, positive, positively, enthusiastic, enthusiasm, playful, playfulness, exciting, excitingly, celebratory, celebratorily, reassuring, reassuringly \\
 & energetic  & vibrant, active, dynamic, energetic, lively, attractive, captivating, stunning, vivid, vibrantly, actively, dynamically, energetically, influential, influence \\
\midrule

\multirow[c]{13}{*}{\rotatebox[origin=c]{90}{\textbf{Neutral}}}
 & anticipated  & likely, probable, expected, projected, forecast, forecasting, potential, predicted, anticipated, anticipation, anticipatory \\
 & transitional & change, changing, shift, shifted, shifting, transition, transitional, transforming, variation, modified, varied, modifying, surprising, surprisingly \\
 & pending      & waiting, pending, postponed, temporary, transient, interim, provisional, paused, delayed, holding \\
 & formal       & official, regulated, regulatory, administrative, controlled, standard, formal, institutional, legal, mandated, officially, formally, objective, objectively \\
 & practical    & economic, financial, commercial, industrial, corporate, market, trade, monetary, fiscal, economically, financially, commercially, expensive, expensively \\
 & statistical  & measured, calculated, statistical, computed, data, metric, empirical, quantified, statistically, empirically, analytical, analytically, factual, factually, comparative, comparatively, insightful, insightfully \\
 & perceived    & observed, perceived, perception, tracked, tracking, monitored, phenomenon, identified, detected, revealing, revealingly \\
 & mixed        & mixed, varied, various, miscellaneous, diverse, multiple, random, differing, diversity, disparate, disparity, divisive, divisively, controversial, controversially \\
 & normal       & normal, normalized, standard, habitual, routine, constant, typical, regular, common, typically, regularly, commonly, neutral, neutrally \\
\midrule

\multirow[c]{22}{*}{\rotatebox[origin=c]{90}{\textbf{Negative}}}
 & declining & decline, declining, decrease, decreased, drop, dropped, falling, shrink, shrinking, downward, declined, decreasing, dropping, fallen, shrunk, disappointing, disappointingly \\
 & crashing  & plummet, plummeted, plunge, plunging, collapse, collapsed, crash, crashing, nosedived, tumble, plummeting, plunged, collapsing, crashed, tumbled, tumbling \\
 & fear      & scared, fear, fears, fearful, frightened, terrified, panic, panicked, apprehensive, dread, scary, fearfully, terrifying, panicking, apprehensively, dreadful, distressing, distressingly, somber, somberly, grave, gravely, anxious, anxiously, tragic, tragically, overwhelming, overwhelmingly \\
 & alarm     & alarm, alarmed, alarming, warning, emergency, threatening, threat, danger, hazardous, alarmingly, warned, threateningly, threatened, dangerous, dangerously, concerning, concerningly, serious, seriously, critical, critically, urgent, urgently, sobering, soberingly, cautious, cautiously, troubling, troublingly, worrisome, worrisomely, disturbing, disturbingly \\
 & anger     & anger, angry, furious, infuriating, outrage, hostile, hostility, resentment, bitter, clashes, angrily, furiously, infuriatingly, outraged, outrageous, resentful, bitterly, clashing, tense, tensely, restrictive, restrictively, frustrating, frustratingly, divided, dividedly \\
 & struggle  & struggle, struggles, struggling, hardship, difficulty, difficulties, tough, challenging, gruelling, struggled, difficult, challenge, problematic, problematically \\
 & pressured & pressure, pressured, pressuring, strained, strains, stretched, burden, burdened, constrained, stretching, burdening, constraining \\
 & harmful   & damage, damaged, damaging, harm, harmful, detrimental, toxic, destructive, devastated, catastrophic, harmfully, detrimentally, toxically, destructively, devastating, catastrophically, disheartening, dishearteningly \\
 & failed    & fail, failed, failing, failure, defeat, defeated, unsuccessful, setback, derailed, mistake, defeating, unsuccessfully, setbacks, mistakes, mistaken \\
 & shortage  & shortage, scarcity, scarce, insufficient, inadequate, lack, missing, deficit, void, empty, shortages, insufficiently, inadequately, lacking, deficits, emptied \\
 & uncertain & uncertain, uncertainty, unclear, unsure, skepticism, dubious, suspicion, ambiguous, tentative, uncertainties, unclearly, skeptical, skeptically, dubiously, suspicious, suspiciously, ambiguously, tentatively \\
 & lonely    & isolated, isolation, abandoned, abandonment, alone, lonely, excluded, alienation, vanished, exclusion, vanishing \\
\bottomrule
\end{tabular}

\end{table}

\newpage
\section{JSON schema}\label{sec:JSON}
The table provided a detailed explanation of the JSON schema adopted in \textit{InfoAffect} dataset for organizing affective annotations. Each key–value pair specified how the detected affects, corresponding confidence scores, and relevant lexical information are stored and accessed. This standardized schema supported transparent data management, reproducibility of experiments, and seamless integration with downstream analytical workflows.
\begin{table}[H]
\centering
\renewcommand{\arraystretch}{2}
\begin{tabular}{lp{0.75\linewidth}}  
\hline
\textbf{Key}              & \textbf{Description}                                                                 \\ \hline
\texttt{affective\_analysis} & The main key containing all affective annotations for the infographic-text pair.      \\ \hline
\texttt{affective\_words}   & An array of words associated with affective expressions found in the text. Each word is accompanied by a confidence score. \\ \hline
\texttt{word}               & Specific affect.                                    \\ \hline
\texttt{confidence}         & The confidence score (0-100). \\ \hline
\end{tabular}
\end{table}

\section{TIDD-EC Prompt Framework}\label{sec:t3}
Table~\ref{tab:t3} presented the detailed configuration of the TIDD-EC prompt framework employed for affect extraction. Each component defined a specific role in guiding multimodal large language models, from specifying the analytical task and procedural instructions to providing explicit behavioral constraints and contextual examples.

\begin{table*}[h]
\caption{Detailed configuration of the TIDD-EC prompt framework for affect extraction.}
\label{tab:t3}
\renewcommand{\arraystretch}{1.5}
\centering
\begin{tabular}{lp{4.5cm}p{10.5cm}}
\hline
Components         & Explanation & Detailed Prompt \textit{in Affect Extraction} \\ 
\hline
T-Task             & Indicate the specific task that MLLMs need to solve. & \textit{Analyze the affective characteristics of the provided infographic (including both visual elements and textual content) and identify affects with confidence scores.} \\
I-Instruction      & Enumerate the steps and instructions that when perform the task. & \textit{Based on the infographics' visual design (colors, layout, typography, imagery) and textual content, identify 5-8 affects that best capture the affective tone. \underline{You must choose only from the provided \textit{Affect Table}.} For each word, assign a confidence score (0-100\%) indicating how strongly this specific affect is perceived.}  \\
D-Do               & Specify the actions that MLLMs should implement during the response process. & \textit{Provide your analysis in the following JSON schema. Output ONLY a single valid JSON object. } \\
D-Don't            & Clarify the behaviors that MLLMs shall refrain from during the response process. & \textit{- Do NOT include any explanations.}
\newline \textit{- Do NOT include any fences. \dots}\\
E-Example          & Present reference cases of desired results or responses that align with expectations. & \textit{- An infographic about climate change with melting ice graphics, alarming statistics, and red color scheme\dots} \\
C-Content          & Provide the relevant data and materials required for task. & \textit{\textgreater\ Infographic Image: [[INSERT IMAGE HERE]]} \newline \textit{\textgreater\ Text Description: [[LEAD TEXT]]}
\newline \textit{\textgreater\ \textit{Affect Table}: [[AFFECT TABLE]]}\\
\hline
\end{tabular}
\end{table*}

\newpage
\section{Participant Information}\label{sec:participants}
Table~\ref{tab:participants} details the demographic and occupational composition of participants who took part in the \textit{InfoAffect} dataset evaluation study. It lists their professional identities, gender, usage scenarios, and age. The group comprises researchers, students, designers, and data professionals with backgrounds in visualization and communication design.

\definecolor{customgreen}{rgb}{0.117, 0.737, 0.239}
\definecolor{customred}{rgb}{0.85,0.13,0.13}
\newcolumntype{C}{>{\centering\arraybackslash}X}
\begin{table*}[h]
\caption{Participants in user study.}
\centering
\renewcommand{\arraystretch}{1.3}
\begin{tabularx}{\textwidth}{@{}c c c C c@{}}
\toprule
\textbf{Participants} & \textbf{Identity} & \textbf{Gender} & \textbf{Usage Scenario} & \textbf{Age} \\
\midrule
E1  & University Professor             & Male   & Academic writing, Visualization teaching                            & 32 \\
E2  & University Lecturer    & Female & Academic writing, Visualization teaching                                                    & 28 \\
P1  & Doctoral student       & Male   & Academic writing, Data analysis                                     & 27 \\
P2  & Master's student       & Male   & Course assignments, Document writing                                & 26 \\
P3  & Doctoral student       & Female & Academic writing, Data analysis                                     & 28 \\
P4  & Data analyst           & Male   & Data analysis, open-data reporting                                  & 33 \\
P5  & Open-data officer      & Male & Policy visualizing, Data analysis              & 30 \\
P6  & Policy researcher      & Female   & Presenting social issues, Data analysis                             & 32 \\
P7  & Data consultant        & Female & Data analysis                                                       & 30 \\
P8 & Marketing analyst      & Male   & Market infographics creating                                         & 31 \\
P9 & Independent designer   & Female & Infographics designer, content publishing                    & 28 \\
P10 & Design blogger         & Male   & Visual designing, blog content editing                              & 27 \\
\bottomrule
\end{tabularx}
\label{tab:participants}
\end{table*}

\newpage
\section{Interview transcripts}\label{sec:interview}
\vspace{4mm}
\newcommand{\Q}[1]{\par\noindent\textbf{Q#1:}~}
\newcommand{\A}[1]{\par\noindent\textbf{A#1:}~}
\noindent
\begin{minipage}[c]{0.42\textwidth}
\textbf{eg.1 E1}\par\vspace{0.8mm}
\includegraphics[width=\linewidth,height=0.3\textheight,keepaspectratio]{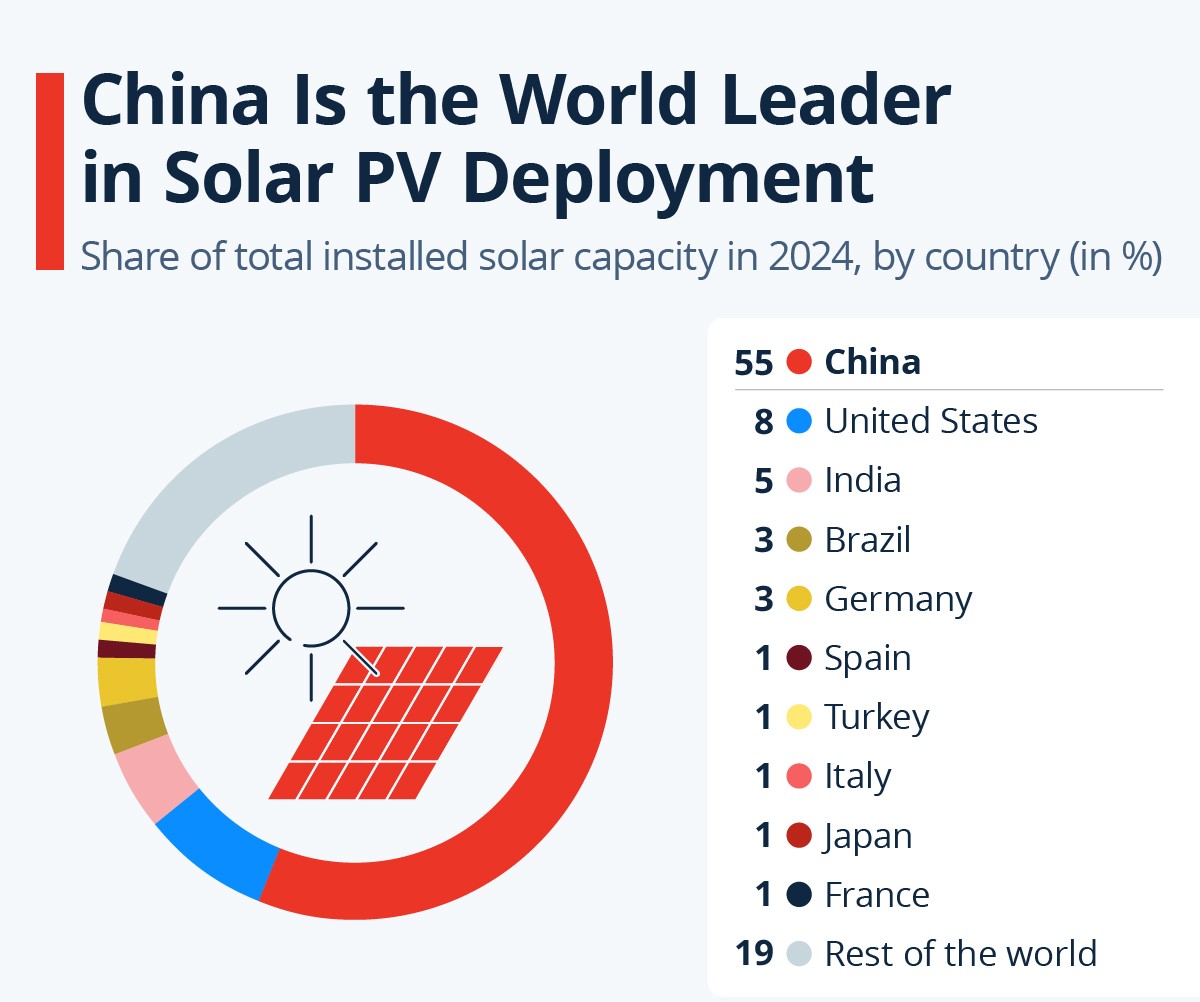}
\end{minipage}\hfill
\begin{minipage}[c]{0.55\textwidth}
\vspace{0pt}
\textit{\textless original intent\textgreater:} I designed this infographic to highlight China’s global leadership in solar photovoltaic deployment using clear quantitative evidence. My primary intention was to convey pride and encouragement toward renewable energy advancement, emphasizing progress and global responsibility rather than competition.\par
\textit{\textless extracted affects\textgreater:} Optimistic, confident, and assertive.

\end{minipage}

\vspace{4mm}

\Q{1}Do you think the affect identified aligns with the affect youoriginally intended to convey in this infographic?
\vspace{0.8mm}
\A{1}Largely, yes. The system's identification of ``optimistic'' and ``confident'' aligns well with my goal of promoting a positive outlook on renewable energy leadership.\\
\Q{2}If the extracted affects differs from your original intent, how would you describe that difference?\vspace{0.8mm}
\A{2}[The extracted affect aligns with the original intent.]
\\
\Q{3}Do you think the visual elements, such as color, composition, or typography, contributed to this unintended affects?\vspace{0.8mm}
\A{3}[The extracted affect aligns with the original intent.]
\\
\Q{4}Does the analysis results help you better understand how affect is represented in your visual design?\vspace{0.8mm}
\A{4}Definitely. It helped me realize how small visual decisions, like using red in a large proportion or positioning the label prominently, can shift the perceived affect from informative to persuasive.
\\
\Q{5}Do you think this kind of affective feedback could be useful for future design processes?\vspace{0.8mm}
\A{5}Yes, very much so. Such feedback would help refine the affective balance in policy communication, ensuring that data visualizations motivate sustainable action without overstating superiority or competitiveness.
\\

\noindent
\begin{minipage}[c]{0.42\textwidth}
\vspace{10mm}
\textbf{eg.2 E2}\par\vspace{0.8mm}
\includegraphics[width=\linewidth,height=0.3\textheight,keepaspectratio]{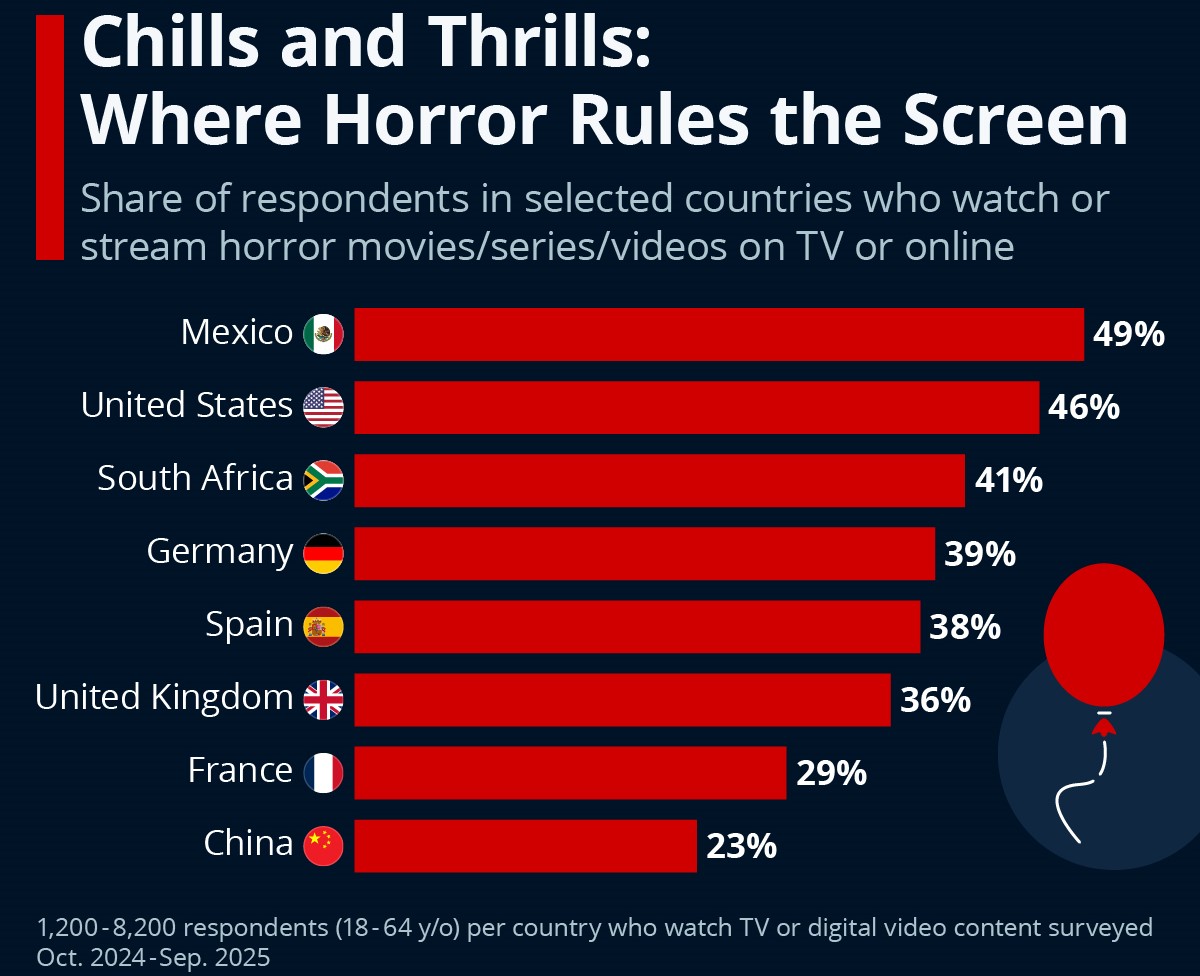}
\end{minipage}\hfill
\begin{minipage}[c]{0.55\textwidth}
\vspace{0pt}
\textit{\textless original intent\textgreater:} I created this infographic to objectively present global viewing preferences for horror movies and series, illustrating which countries show the highest engagement with this genre. My intent was primarily informative rather than affective, aiming to highlight a cultural trend through clean visual comparison.\par
\textit{\textless extracted affects\textgreater:} Fearful, intense.
\end{minipage}

\vspace{4mm}
\Q{1}Do you think the affect identified aligns with the affect you originally intended to convey in this infographic?
\vspace{0.8mm}
\A{1}Not exactly. The system saw it as ``sinister'' or ``fearful'' but I only meant to report viewing data with subtle horror-themed coherence, not create an unsettling vibe.\\

\Q{2}If the extracted affects differs from your original intent, how would you describe that difference?
\vspace{0.8mm}
\A{2}The system exaggerated the tone. I wanted mild thematic links to horror, not to directly evoke fear as it interpreted.\\

\Q{3}Do you think the visual elements, such as color, composition, or typography, contributed to this unintended affects?
\vspace{0.8mm}
\A{3}Probably. The dark background, bright red bars, and red balloon likely amplified fear associations, making it more disturbing than intended.\\

\Q{4}Does the analysis results help you better understand how affect is represented in your visual design?
\vspace{0.8mm}
\A{4}Yes. Even small stylistic choices can strongly shape affective interpretation, despite neutral data.\\

\Q{5}Do you think this kind of affective feedback could be useful for future design processes?
\vspace{0.8mm}
\A{5}Absolutely. It'll help me balance ``genre flavor'' and avoid unintended affective responses while keeping thematic relevance.\\

\noindent
\begin{minipage}[c]{0.42\textwidth}
\vspace{10mm}
\textbf{eg.3 P2}\par\vspace{0.8mm}
\includegraphics[width=\linewidth,height=0.3\textheight,keepaspectratio]{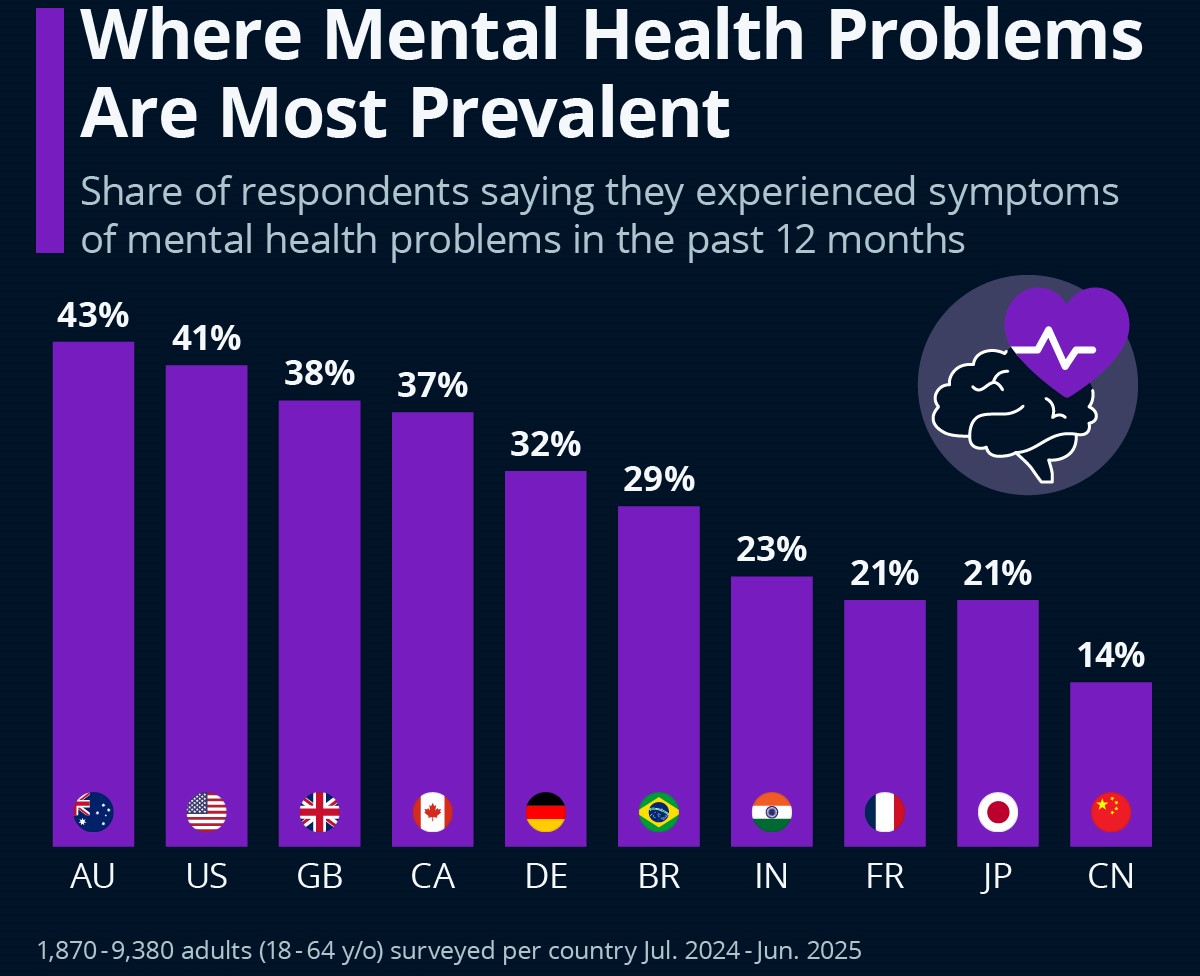}
\end{minipage}\hfill
\begin{minipage}[c]{0.55\textwidth}
\vspace{0pt}
\textit{\textless original intent\textgreater:} I designed this infographic to show global differences in reported mental health symptoms in a clear and empathetic way. My main goal was to raise awareness, not to dramatize the data.\par
\textit{\textless extracted affects\textgreater:} Calm, empathetic, serious.
\end{minipage}

\vspace{4mm}

\Q{1}Do you think the affect identified aligns with the affect you originally intended to convey in this infographic?
\vspace{0.8mm}
\A{1}Yes, completely. The system picked up on exactly what I wanted, a calm and empathetic tone that shows the data respectfully.\\
\Q{2}If the extracted affects differs from your original intent, how would you describe that difference?
\vspace{0.8mm}
\A{2}[The extracted affect aligns with the original intent.]\\
\Q{3}Do you think the visual elements, such as color, composition, or typography, contributed to this unintended affects?
\vspace{0.8mm}
\A{3}[The extracted affect aligns with the original intent.]\\
\Q{4}Does the analysis results help you better understand how affect is represented in your visual design?
\vspace{0.8mm}
\A{4}Yes, it confirms that even simple design choices, like color and icon style, can really shape how people affectively read the data.\\
\Q{5}Do you think this kind of affective feedback could be usefulfor future design processes?
\vspace{0.8mm}
\A{5}For sure. Getting this kind of feedback helps me make sure I'm hitting the right affective tone, especially for sensitive topics like mental health.\\
\noindent
\begin{minipage}[c]{0.42\textwidth}
 \vspace{10mm}
\textbf{eg.4 P3}\par\vspace{0.8mm}
\includegraphics[width=\linewidth,height=0.3\textheight,keepaspectratio]{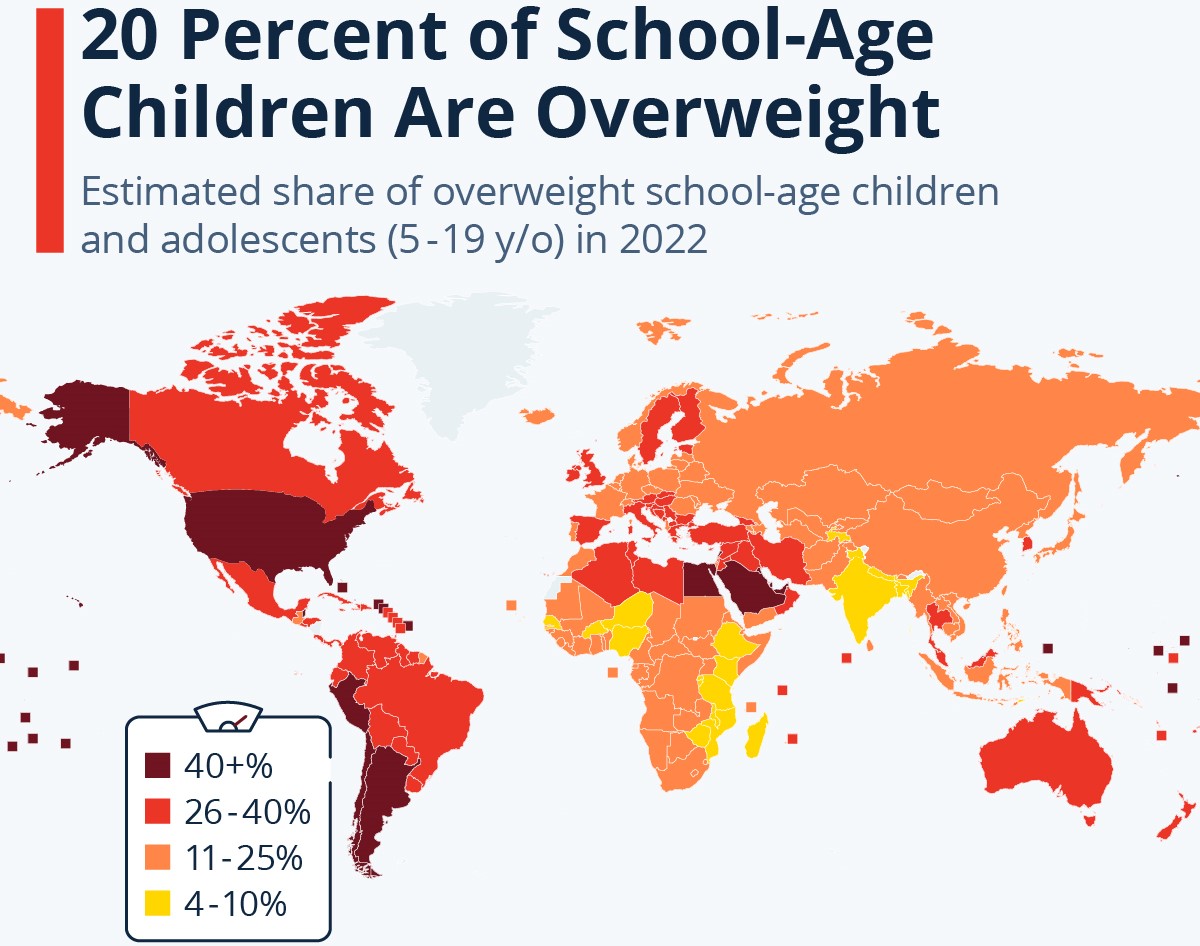}
\end{minipage}\hfill
\begin{minipage}[c]{0.55\textwidth}
\vspace{0pt}
\textit{\textless original intent\textgreater:} I made this infographic simply to list the relevant figures in my manuscript, keeping the tone descriptive rather than evaluative.\par
\textit{\textless extracted affects\textgreater:} alarming, concerning.
\end{minipage}

\vspace{4mm}
\Q{1}Do you think the affect identified aligns with the affect you originally intended to convey in this infographic?
\vspace{0.8mm}
\A{1}No. My original intent was to keep the tone descriptive rather than evaluative, but the identified affect was negative.\\

\Q{2}If the extracted affects differs from your original intent, how would you describe that difference?
\vspace{0.8mm}
\A{2}I intended to simply list the relevant figures, but the detected affect appeared ``alarming'' and ``concerning'' which differs from her neutral, factual intent.\\

\Q{3}Do you think the visual elements, such as color, composition, or typography, contributed to this unintended affects?
\vspace{0.8mm}
\A{3}Yes. I reflect that it's due to the warm red--orange ramp and the strong contrast in the legend.\\

\Q{4}Does the analysis results help you better understand how affect is represented in your visual design?
\vspace{0.8mm}
\A{4}Yes. It helps me realize how color and contrast can unintentionally convey negative affectives even in descriptive visualizations.\\

\Q{5}Do you think this kind of affective feedback could be usefulfor future design processes?
\vspace{0.8mm}
\A{5}Yes. Such feedback could help designers anticipate and adjust unintended affective tones in their visual work.\\

\noindent
\begin{minipage}[c]{0.42\textwidth}
\textbf{eg.5 P4}\par\vspace{0.8mm}
\includegraphics[width=\linewidth,height=0.3\textheight,keepaspectratio]{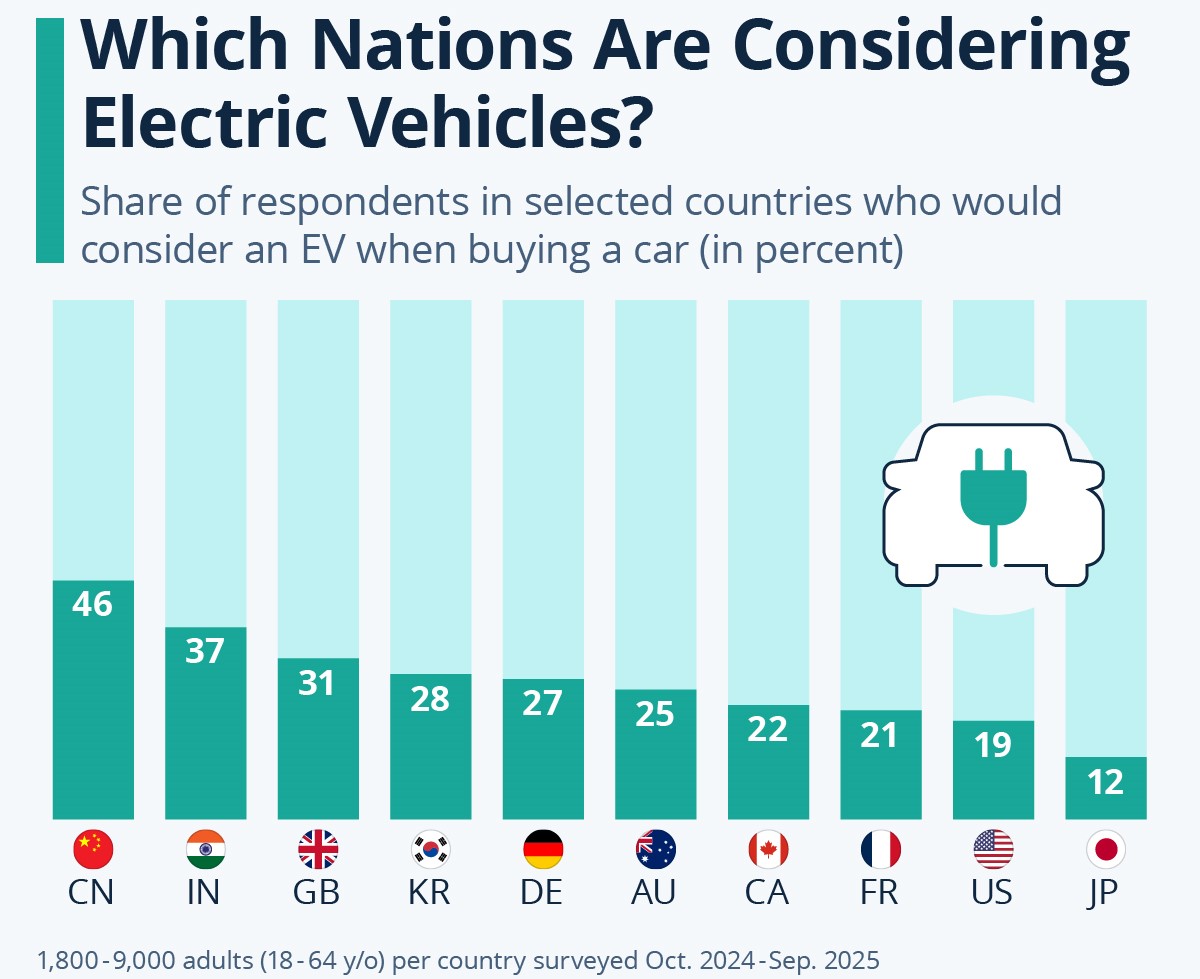}
\end{minipage}\hfill
\begin{minipage}[c]{0.55\textwidth}
\vspace{0pt}
\textit{\textless original intent\textgreater:} I submitted this infographic aimed to naturally present the information, simply enumerated which countries are more open to electric vehicles without taking an evaluative stance.\par
\textit{\textless extracted affects\textgreater:} informational.
\end{minipage}

\vspace{4mm}

\Q{1}Do you think the affect identified aligns with the affect you originally intended to convey in this infographic?
\vspace{0.8mm}
\A{1}Yes, perfectly. My goal was just to report which nations are considering electric vehicles, the analysis produced neutral terms that align with the report-style tone I intended.\\
\Q{2}If the extracted affects differs from your original intent, how would you describe that difference?
\vspace{0.8mm}
\A{2}[The extracted affect aligns with the original intent.]\\
\Q{3}Do you think the visual elements, such as color, composition, or typography, contributed to this unintended affects?
\vspace{0.8mm}
\A{3}[The extracted affect aligns with the original intent.]\\
\Q{4}Does the analysis results help you better understand how affect is represented in your visual design?
\vspace{0.8mm}
\A{4}Yes. It confirms that using simple bar graphs, neutral colors, and clear labels can effectively convey an informational tone, helping me grasp how visual choices shape affect representation.\\
\Q{5}Do you think this kind of affective feedback could be usefulfor future design processes?
\vspace{0.8mm}
\A{5}For sure. It ensures I maintain the neutral, informative tone I want.\\

\noindent
\begin{minipage}[c]{0.42\textwidth}
\textbf{eg.6 P5}\par\vspace{0.8mm}
\includegraphics[width=\linewidth,height=0.3\textheight,keepaspectratio]{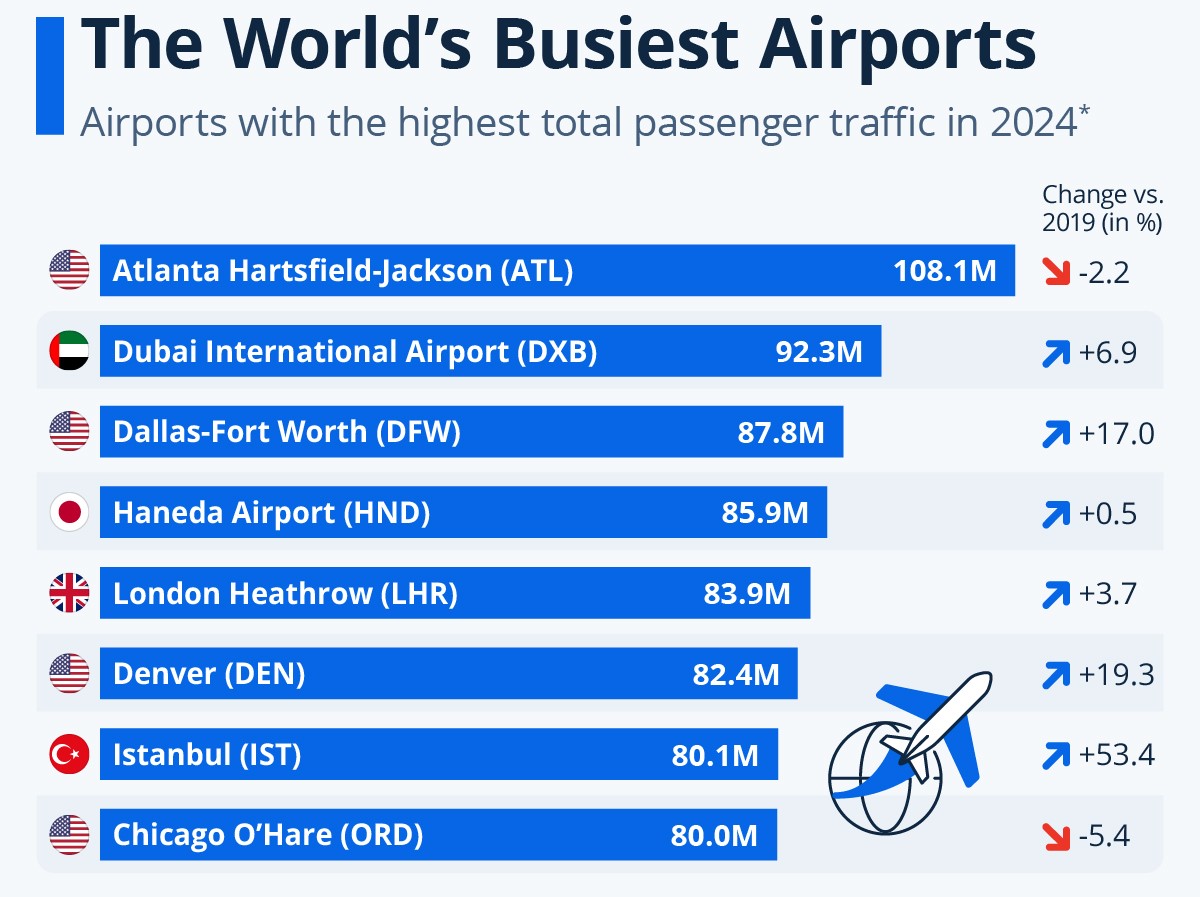}
\end{minipage}\hfill
\begin{minipage}[c]{0.55\textwidth}
\vspace{0pt}
\textit{\textless original intent\textgreater:} I made this infographic to simply report the world’s busiest airports’ passenger traffic data, aiming for a neutral, factual tone.\par
\textit{\textless extracted affects\textgreater:} Informational, neutral.
\end{minipage}

\vspace{4mm}
\Q{1}Do you think the affect identified aligns with the affect you originally intended to convey in this infographic?
\vspace{0.8mm}
\A{1}Yes, totally. I just wanted to lay out the airport traffic facts, and the system picked up ``informational, neutral'' vibes, exactly what I intended.\\

\Q{2}If the extracted affects differs from your original intent, how would you describe that difference?
\vspace{0.8mm}
\A{2}[The extracted affect aligns with the original intent.]\\

\Q{3}Do you think the visual elements, such as color, composition, or typography, contributed to this unintended affects?
\vspace{0.8mm}
\A{3}[The extracted affect aligns with the original intent.]\\

\Q{4}Does the analysis results help you better understand how affect is represented in your visual design?
\vspace{0.8mm}
\A{4}Definitely. It shows that using straightforward visuals and a neutral color scheme can nail an informational tone, so I get how design choices shape that now.\\

\Q{5}Do you think this kind of affective feedback could be usefulfor future design processes?
\vspace{0.8mm}
\A{5}For sure. It helps me make sure I keep that neutral, data-driven tone in future infographics where I just want to present facts.\\

\noindent
\begin{minipage}[c]{0.42\textwidth}
\textbf{eg.7 P6}\par\vspace{0.8mm}
\includegraphics[width=\linewidth,height=0.3\textheight,keepaspectratio]{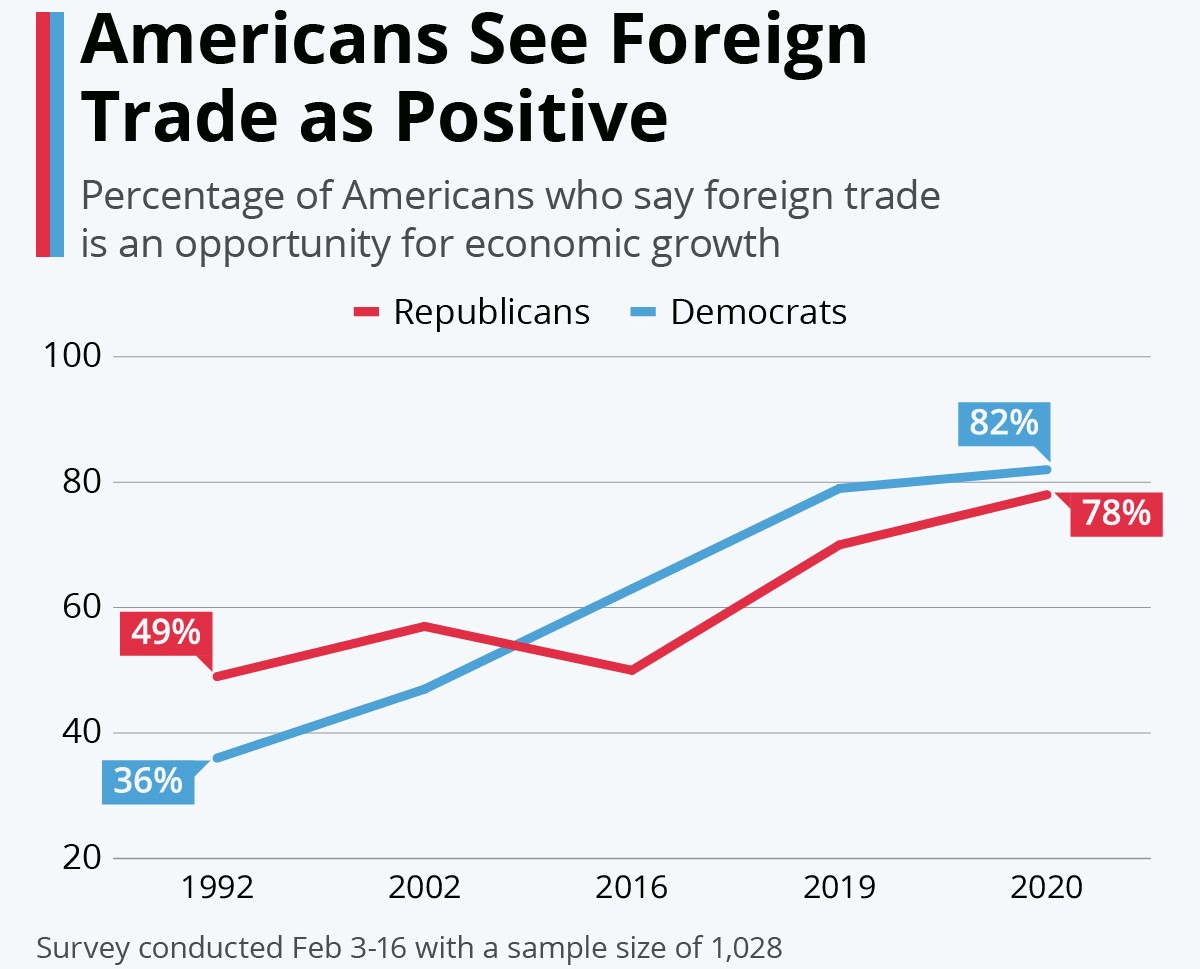}
\end{minipage}\hfill
\begin{minipage}[c]{0.55\textwidth}
\vspace{0pt}
\textit{\textless original intent\textgreater:} I offered this infographic aiming for “hopeful”, seeking a positive affect.\par
\textit{\textless extracted affects\textgreater:} optimistic.
\end{minipage}

\vspace{4mm}

\Q{1}Do you think the affect identified aligns with the affect you originally intended to convey in this infographic?
\vspace{0.8mm}
\A{1}Yes. My intent in the policy report was to highlight a hopeful trend for policymakers’ consideration, and the extracted affect of the infographic was ``optimistic'' which closely aligned with my original intent.\\
\Q{2}If the extracted affects differs from your original intent, how would you describe that difference?
\vspace{0.8mm}
\A{2}[The extracted affect aligns with the original intent.]\\
\Q{3}Do you think the visual elements, such as color, composition, or typography, contributed to this unintended affects?
\vspace{0.8mm}
\A{3}[The extracted affect aligns with the original intent.]\\
\Q{4}Does the analysis results help you better understand how affect is represented in your visual design?
\vspace{0.8mm}
\A{4}Definitely. It confirms that using upward rending lines and distinct colors can effectively convey an optimistic, hopeful tone, deepening my understanding of how visual elements shape affect representation.\\
\Q{5}Do you think this kind of affective feedback could be usefulfor future design processes?
\vspace{0.8mm}
\A{5}For sure. It helps me ensure that the optimistic, hopeful tone I intend is consistently conveyed in future designs, especially when highlighting positive trends like this one.\\

\noindent
\begin{minipage}[c]{0.42\textwidth}
\textbf{eg.8 P7}\par\vspace{0.8mm}
\includegraphics[width=\linewidth,height=0.3\textheight,keepaspectratio]{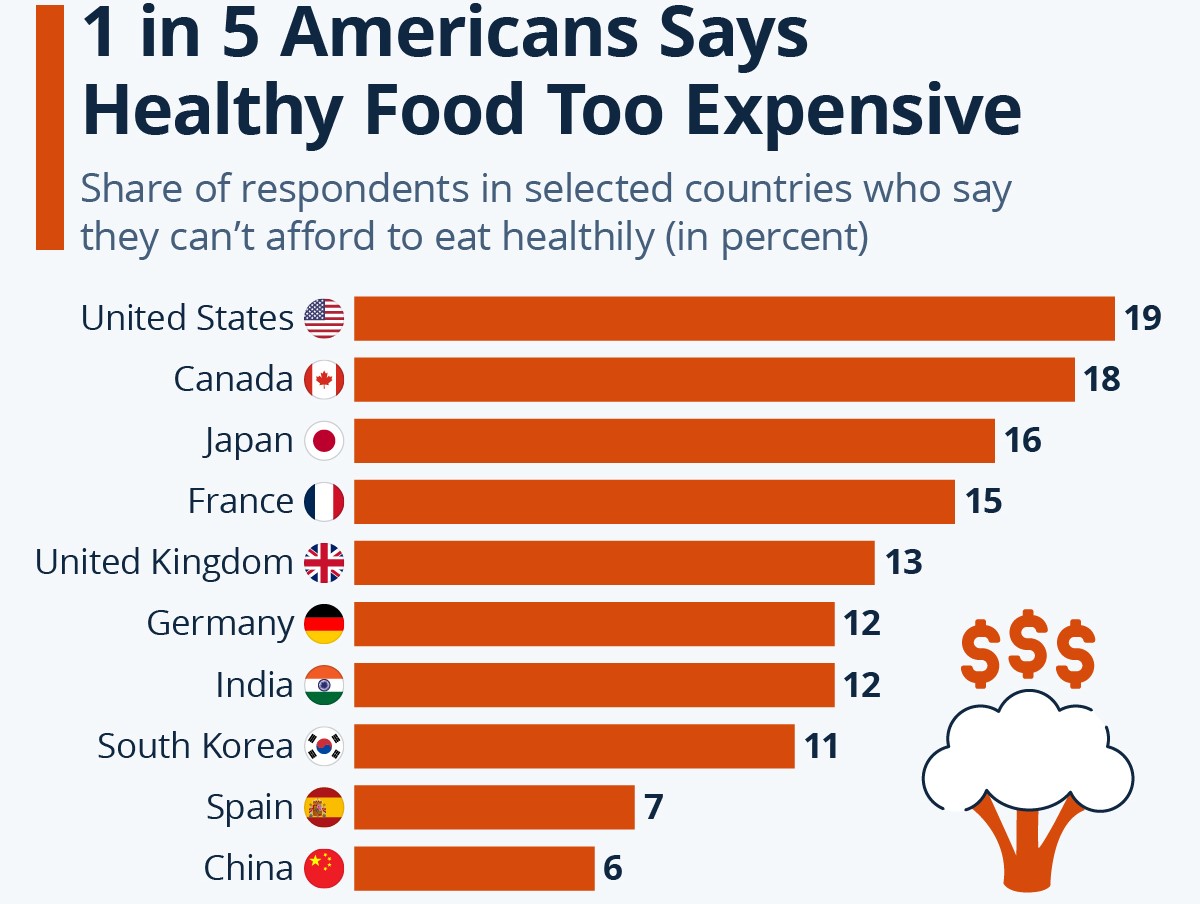}
\end{minipage}\hfill
\begin{minipage}[c]{0.55\textwidth}
\vspace{0pt}
\textit{\textless original intent\textgreater:} I created this infographic to show the global situation of people finding healthy food too expensive, aiming to present the data in a somewhat approachable and relatable way.\par
\textit{\textless extracted affects\textgreater:} Informative, concerned.
\end{minipage}

\vspace{4mm}
\Q{1}Do you think the affect identified aligns with the affect you originally intended to convey in this infographic?
\vspace{0.8mm}
\A{1}Not really. I wanted it to be more approachable, but the system picked up an informative and concerned tone.\\

\Q{2}If the extracted affects differs from your original intent, how would you describe that difference?
\vspace{0.8mm}
\A{2}My intent was to make it relatable, but the system's identified affect is more focused on being informative and concerned, missing the intended approachable vibe.\\

\Q{3}Do you think the visual elements, such as color, composition, or typography, contributed to this unintended affects?
\vspace{0.8mm}
\A{3}Yeah. The orange bar graph, country flags, and the serious ``\$\$\$'' and broccoli icon make it feel informative and concerned, which is not as approachable as he wanted.\\

\Q{4}Does the analysis results help you better understand how affect is represented in your visual design?
\vspace{0.8mm}
\A{4}Definitely. It shows that even if you want an approachable tone, the visual choices like bar style and icons can push it toward a more informative and concerned affect.\\

\noindent
\begin{minipage}[c]{0.42\textwidth}
\textbf{eg.9 P10}\par\vspace{0.8mm}
\includegraphics[width=\linewidth,height=0.3\textheight,keepaspectratio]{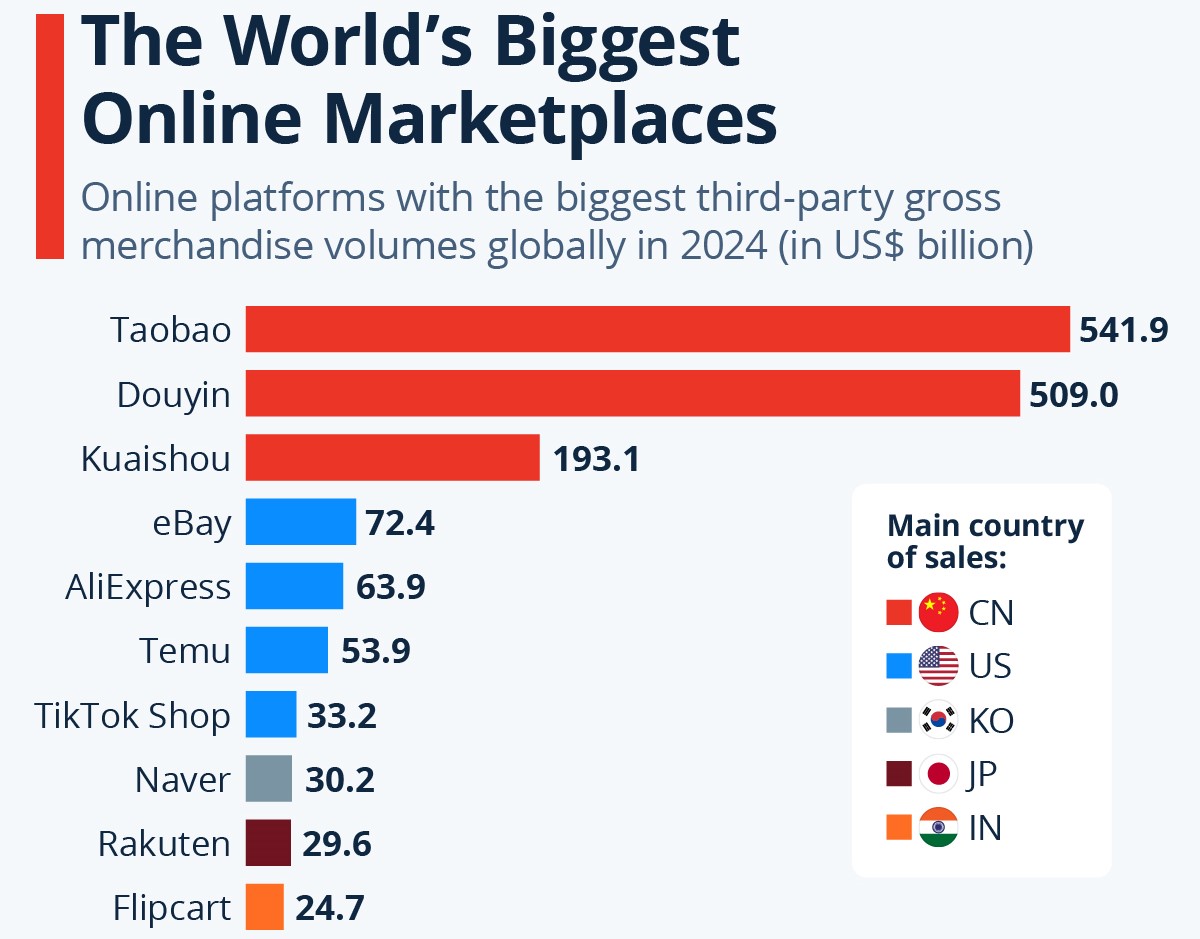}
\end{minipage}\hfill
\begin{minipage}[c]{0.55\textwidth}
\vspace{0pt}
\textit{\textless original intent\textgreater:} I made this infographic to express a negative winner-takes-all trend, namely a worrying concentration in which one platform dominates the others.\par
\textit{\textless extracted affects\textgreater:} confident, competitive.
\end{minipage}

\vspace{4mm}

\Q{1}Do you think the affect identified aligns with the affect you originally intended to convey in this infographic?
\vspace{0.8mm}
\A{1}No. My intent is to express a negative winner-takes-all trend, but the identified affect was positive.\\
\Q{2}If the extracted affects differs from your original intent, how would you describe that difference?
\vspace{0.8mm}
\A{2}I want to show “a worrying concentration in which one platform dominates'' yet the extracted affects are ``confident'' and ``competitive'' giving a sense of achievement rather than imbalance.\\
\Q{3}Do you think the visual elements, such as color, composition, or typography, contributed to this unintended affects?
\vspace{0.8mm}
\A{3}Yes. I reflected that the bright, saturated palette and clean ranking layout make it feel like achievement rather than imbalance, and the long red top bar reads as ``leading'' more than ``concerning'', while the crisp flags and labels add a celebratory tone.\\
\Q{4}Does the analysis results help you better understand how affect is represented in your visual design?
\vspace{0.8mm}
\A{4}Yes. It reveals how design choices such as palette, layout, and iconography can shift the affective reading from critical to celebratory.\\
\Q{5}Do you think this kind of affective feedback could be usefulfor future design processes?
\vspace{0.8mm}
\A{5}For sure. Such feedback can help designers better align visual tone with their intended message and avoid unintended positive framing of negative topics.\\

\Q{5}Do you think this kind of affective feedback could be usefulfor future design processes?
\vspace{0.8mm}
\A{5}For sure. It helps me see where his design choices don't match my intended tone, so he can adjust things like visuals or color to get the affect he wants next time.\\

\end{document}